\documentclass[journal]{IEEEtran}
\ifCLASSINFOpdf
\else
\fi
\usepackage{xcolor,soul,framed} 

\colorlet{shadecolor}{yellow}

\usepackage[cmex10]{amsmath}
\usepackage{algorithm}
\usepackage{algorithmic}
\usepackage{array}
\usepackage{mdwmath}
\usepackage{mdwtab}
\usepackage{eqparbox}
\usepackage{url}
\usepackage{amsmath}

\usepackage[pdftex]{graphicx}
\usepackage{subfigure}
\graphicspath{ {./} }

\usepackage{booktabs} 
\usepackage{multirow} 
\usepackage{amssymb}
\usepackage{bbding}

\begin{document}
%
\title{MoCom: Motion-based Inter-MAV Visual Communication Using Event Vision and \\Spiking Neural Networks}
%
%
%

\author{Zhang Nengbo$^{1}$,
     Hann~Woei~Ho$^{1,*}$,~\IEEEmembership{Member,~IEEE,}
         Ye~Zhou$^{1}$,~\IEEEmembership{Member,~IEEE,}
 \thanks{$^{*}$Corresponding author.
 }
 \thanks{$^{1}$Zhang Nengbo, Hann Woei Ho, and Ye Zhou are with School of Aerospace Engineering, Engineering Campus, Universiti Sains Malaysia, 14300 Nibong Tebal, Pulau Pinang, Malaysia (email: zhangnb@student.usm.my; aehannwoei@usm.my; zhouye@usm.my).
 }
 }

%
%

\markboth{IEEE Transactions on Robotics, Submitted}%
{Shell \MakeLowercase{\textit{et al.}}: Bare Demo of IEEEtran.cls for IEEE Journals}
%



\maketitle

\begin{abstract}

Reliable communication in Micro Air Vehicle (MAV) swarms is challenging in environments, where conventional radio-based methods suffer from spectrum congestion, jamming, and high power consumption. Inspired by the waggle dance of honeybees, which efficiently communicate the location of food sources without sound or contact, we propose a novel visual communication framework for MAV swarms using motion-based signaling. In this framework, MAVs convey information, such as heading and distance, through deliberate flight patterns, which are passively captured by event cameras and interpreted using a predefined visual codebook of four motion primitives: vertical (up/down), horizontal (left/right), left-to-up-to-right, and left-to-down-to-right, representing control symbols (``start'', ``end'', ``1'', ``0''). To decode these signals, we design an event frame-based segmentation model and a lightweight Spiking Neural Network (SNN) for action recognition. An integrated decoding algorithm then combines segmentation and classification to robustly interpret MAV motion sequences. Experimental results validate the framework's effectiveness, which demonstrates accurate decoding and low power consumption, and highlights its potential as an energy-efficient alternative for MAV communication in constrained environments.
\end{abstract}


%
\IEEEpeerreviewmaketitle

\section{Introduction}
\begin{figure}
\centering 
\includegraphics[width=0.48\textwidth]{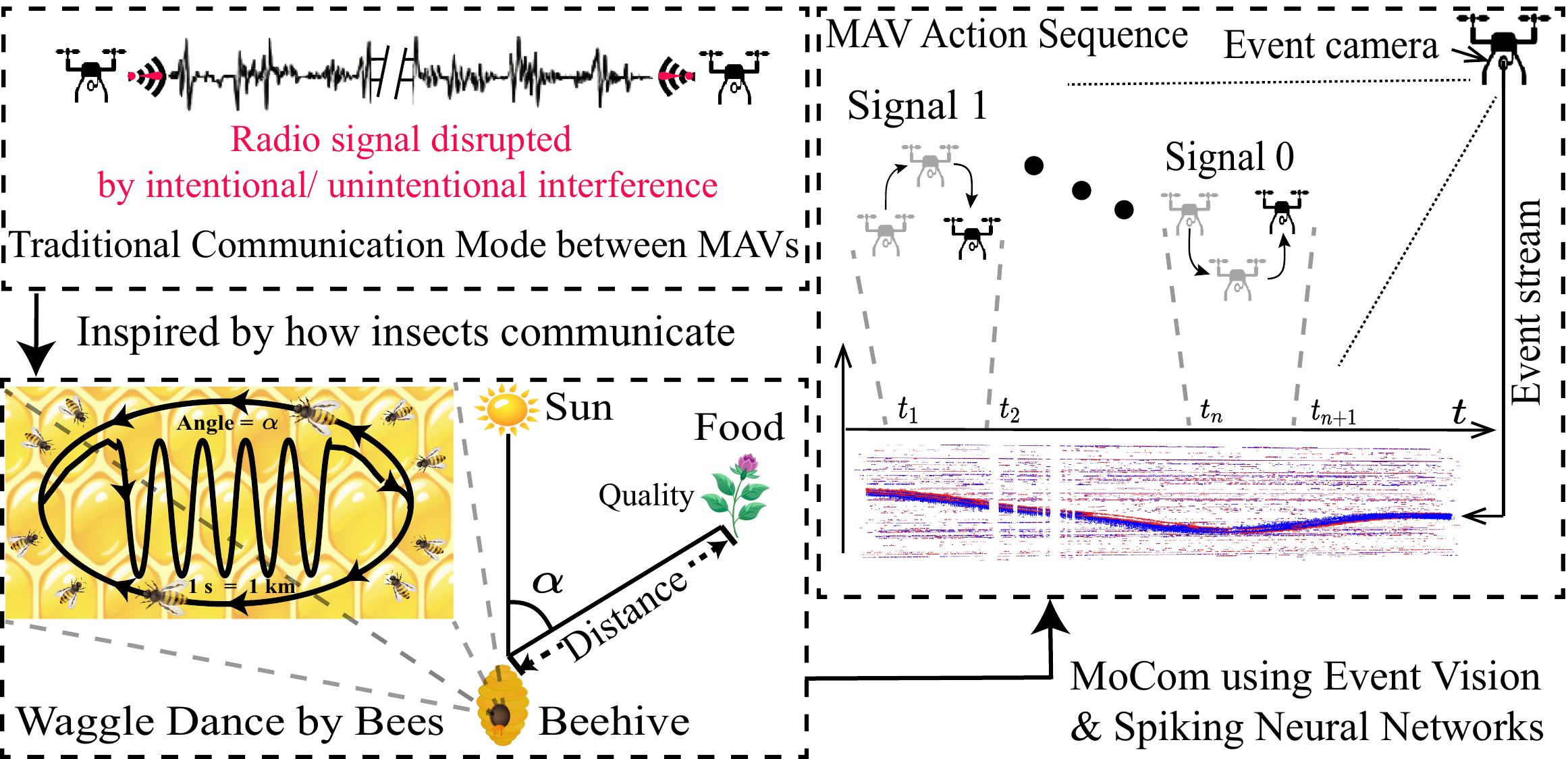}
\caption{Overview of the proposed motion-based inter-MAV visual communication (MoCom) framework for MAV swarms. Traditional radio-based communication can be easily disrupted in constrained environments. Inspired by the waggle dance of honeybees, a motion-based signaling approach, where MAVs convey information through deliberate flight patterns. These motions are captured by event cameras. The event stream is then processed using a biologically inspired pipeline that segments the motion and classifies it using a spiking neural network. This paradigm offers a decentralized and interference-resilient alternative to conventional communication in MAV swarms.} 
\label{fig:bee}
\end{figure}

\IEEEPARstart{I}{n} nature, many species have evolved non-verbal communication strategies, such as the waggle dance of honeybees, pheromone trails in ants, and postural displays in wolves, to support social coordination and collective intelligence. Among them, the waggle dance of honeybees is a remarkable example (see Fig. \ref{fig:bee}). Through specific body movements, bees encode and transmit information about food location, direction, and distance, as studied in a work \cite{dong2023social}. This can enable the colony to make distributed foraging decisions without centralized control. This biologically evolved action-based communication system equips honeybee swarms with self-organization, robustness to environmental noise, and energy efficiency, offering valuable inspiration for engineered multi-agent systems \cite{su2024fixed}. 

Such bio-inspired communication paradigms present compelling solutions for modern Micro Air Vehicle (MAV) swarms, which face critical challenges in maintaining reliable and covert inter-agent coordination. Particularly in high-stakes applications, including disaster rescue  and environmental monitoring, conventional wireless protocols, such as Wi-Fi, radio, 4G/5G, suffer from spectrum scarcity \cite{fantacci2022multi}, jamming vulnerabilities \cite{pirayesh2022jamming}, and poor scalability \cite{zhou2023toward}, which can lead to communication failures in dense MAV swarms. These limitations become debilitating in contested or GPS-denied environments, where resilience and stealth are paramount. Notably, the honeybee’s motion-encoded communication model aligns precisely with MAV swarms’ need for decentralized, interference-resistant, and energy-efficient alternatives.

Inspired by the decentralized and robust communication of honeybees, which encodes and delivers information through physical movements, we propose a novel ``action as signal'' paradigm for MAV swarms to address the limitations of conventional wireless communication. In this paradigm, MAVs encode information in deliberate motion patterns, much like bees conveying foraging details through physical movements. Such motion-based communication not only offers a decentralized, interference-resistant, and energy-efficient alternative to radio-based protocols but also naturally aligns with the perception-action loop of autonomous systems. This synergy enables integration in dynamic, unstructured environments.
To capture motion-based signals effectively, a sensing modality must meet three requirements: high temporal resolution, robustness to lighting variations, and sensitivity to rapid movements. Here, event cameras emerge as an ideal solution. Unlike conventional frame-based RGB cameras, they asynchronously detect pixel-level brightness changes, offering low latency, high dynamic range, and sparse data output. These capabilities are critical for reliably decoding the fast, edge-rich motion patterns generated by MAVs, while maintaining energy efficiency, a feature particularly vital for MAV swarm operations. 
Recent advances have demonstrated the potential of event cameras in MAV behavior recognition, such as object tracking \cite{xue2023smalltrack} and pose estimation \cite{hanyu2024absolute}. However, most of these applications focus on short-duration event segments and do not address the challenges of processing long, continuous event streams typical in real-time MAV communication. Existing MAV communication methods still largely rely on RF-based techniques, which often fail in environments with electromagnetic interference or obstructed lines of sight, leading to unreliable connections and potential mission failure. Furthermore, the ability to extract consistent, interpretable signals from noisy event streams during dynamic flight remains an open challenge, underscoring the need for a new communication framework tailored to the unique characteristics of event-based vision.

Event data, despite its sparsity and asynchronous nature, contains rich temporal structure. Frame-wise statistical indicators, such as the total number of events and the ratio of positive to negative events, naturally encode motion dynamics, providing valuable cues for segmenting continuous event streams into discrete action units. These features offer a lightweight, low-computation, and intuitive foundation for temporal analysis, aligning seamlessly with the properties of event-based sensing. As such, statistical segmentation of event data is not only effective but also particularly suitable for real-time MAV applications with constrained computational resources.

To build a fully bio-inspired MAV visual communication model, we further incorporate a brain-inspired biometric classifier based on Spiking Neural Networks (SNNs) \cite{tavanaei2019deep}. SNNs process information through discrete spikes and inherently support event-driven computation, making them highly compatible with the output of event cameras. Compared to traditional neural architectures, such as Convolutional Neural Networks (CNNs) or Recurrent Neural Networks (RNNs), SNNs offer improved energy efficiency and better handling of sparse, asynchronous inputs. In our system, a lightweight SNN model is employed to classify segmented MAV motion clips into predefined communication symbols (e.g., binary digits), enabling end-to-end decoding of visual signals in a bio-inspired manner.

\begin{figure*}[htbp]
		\centering 
		\includegraphics[width=0.8\textwidth]{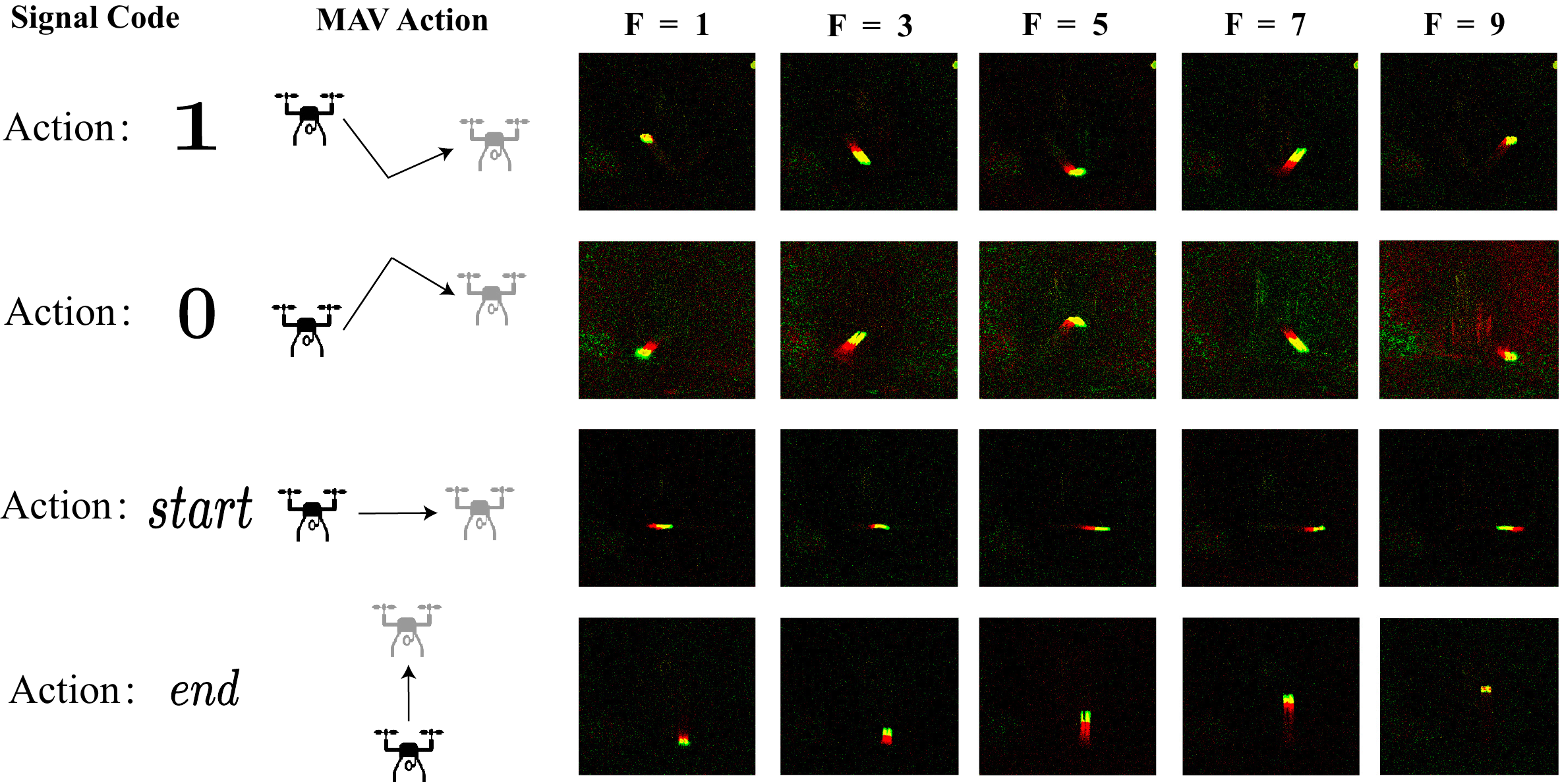}
            
		\caption{Several common visual communication codes are presented. Specifically, we define four types of communication codes (action : ``start'',  action : ``end'',  action : ``1'', action : ``0'').  A single event MAV motion is segmented into $F$-th event block based on an equal event count strategy, and the accumulated events are used to construct $F$ event frames. Red color represents positive events, green color represents negative events, and the yellow regions indicate the overlap between positive and negative events.} 
        \label{fig:twoCamera}
\end{figure*}

In this paper, we propose an event vision-based framework for inter-MAV message transmission. It accurately recognizes MAV actions and decodes action sequences into meaningful messages, enabling visual communication between MAVs. The defined visual communication codes are illustrated in Fig. \ref{fig:twoCamera}. To the best of our knowledge, this is the first work to introduce the concept of motion-based visual communication into MAV systems. The main contributions of this paper are threefold:
\begin{itemize}
  \item  design a \textit{MAV motion segmentation model} to process long continuous event streams by segmenting MAV motion-encoded signals into discrete semantic units. This facilitates the subsequent decoding of MAV motion information. 
  \item develop a lightweight yet accurate \textit{MAV motion recognition model} (EventMAVNet) inspired by biological neural processing, utilizing a spiking neural network architecture. Compared to other SNN recognition models, we achieve both computational efficiency and high recognition precision.
  \item introduce the \textit{Integrated MAV Segmentation and Recognition (IMSR) algorithm}, a MAV motion sequence decoding approach that seamlessly integrates event signals segmentation and recognition. The IMSR algorithm integrates segmentation and recognition to robustly decode MAV motion sequence signals, effectively mitigating noise and interference in dynamic environments.
\end{itemize}

The rest of this paper is organized as follows: Section 2 reviews related works on MAV communication and recent advancements in event recognition models based on spiking neural networks. Section 3 details our proposed method, including the event signal segmentation model, motion recognition model, and MAV motion signal decoding algorithms. Section 4 presents the experimental results and performance evaluation. Section 5  concludes the paper and details future directions for inter-MAV visual communication. Through this work, we not only advance the application of event cameras in MAV signal transmission but also provide new insights into integrating vision and communication in dynamic environments. 

\section{Related Works }
This section reviews key research in MAV communication techniques, event-based MAV action recognition, and motion segmentation using event frames.

\subsection{MAV communication techniques}

MAV communication technology \cite{javaid2023communication} is a key component in MAV swarms \cite{javed2024state}, enabling swarm formation and coordination for applications, such as search and rescue and temporary communication networks \cite{calvo2024networked} in disaster-affected areas. Generally, MAV communication can be classified into distributed \cite{montijano2013distributed,gao2022coverage, talamali2021less} and centralized approaches \cite{hashim2024advances,liang2019reconfigurable,khawaja2019survey,hashim2021gps}. Regarding distributed MAV communication, an early study \cite{montijano2013distributed} introduced a distributed information exchange method based on neighborhood data sharing to address data association problem in multi-robot systems. Subsequent studies \cite{gao2022coverage, talamali2021less} have demonstrated that decentralized coordination enhances swarm control and adaptability. Similarly, distributed communication principles have been successfully applied to vision-based multi-MAV control \cite{aranda2015formation}. Decentralized communication phenomena are also observed in natural swarm organizations, such as social insect colonies \cite{dong2023social}. In contrast, centralized MAV communication typically adopts a star topology, where all MAVs communicate through a central master node. 
This approach relies on MAV onboard electronics \cite{hashim2024advances} and uses methods like point-to-point \cite{liang2019reconfigurable}, ground-to-air \cite{khawaja2019survey}, and GPS-based communication \cite{hashim2021gps}. These methods ensure robust swarm control and coordination.

Traditional approaches, while effective, face challenges such as spectrum scarcity and vulnerability to interference, prompting recent advancements in novel MAV communication techniques. For example, one study \cite{yang2019power} explored Visible Light Communication (VLC) to simultaneously enable communication and illumination. Another work \cite{schelle2019visual} utilized vision-based analysis for direct MAV-operator interaction. However, these communication technologies all use visual light as a carrier for information transmission, and do not use the MAV's motion pattern to transmit information. Additionally, emerging research \cite{meng2023uav, mu2023uav} suggests that future MAV communication will increasingly integrate with sensing technologies, leading to unified communication-perception MAV systems.

\subsection {Action Recognition in Event Vision Using Spiking Neural Networks}
Action recognition is a classic pattern classification task. Traditional action recognition methods primarily focus on video data, targeting human actions, and employ deep learning models combined with temporal modeling techniques, such as optical flow-based Two-Stream networks \cite{feichtenhofer2016convolutional}, Long Short Term Memory networks (LSTM) \cite{majd2020correlational}, and more recently, Transformer architectures \cite{xing2023svformer}. However, these approaches are mostly designed for RGB video inputs and are not directly applicable to event camera data.   

For event-based action recognition, researchers have proposed various event data representations, such as event frame accumulation \cite{innocenti2021temporal}, voxel grids \cite{xie2024event}, and time surfaces \cite{sironi2018hats}. Nevertheless, these representations often lead to the loss of the inherent sparsity and asynchronous nature of event streams, limiting the advantages of event cameras.
To address these challenges, Spiking Neural Networks (SNNs) \cite{maass1997networks} have emerged as a promising alternative for processing event-based data \cite{sekikawa2019eventnet}. Unlike conventional artificial neural networks, SNNs naturally handle sparse and asynchronous information, making them well-suited for event-based motion analysis. However, due to their spike-based discrete computation and non-differentiable activation functions, SNNs suffer from challenges in gradient-based optimization, leading to difficulties in achieving high-performance training through conventional backpropagation. This limitation has hindered their widespread adoption in action recognition tasks. To address this, recently, there are numerous works \cite{wu2018spatio,zheng2021going,kim2021n,kim2022neural} focusing on directly training high performance spiking neural networks. Initially, to address the challenge of non-differentiability in SNNs during training, the Spatio-Temporal Back Propagation (STBP) algorithm \cite{wu2018spatio} was proposed. By simultaneously considering the layer-by-layer Spatial Domain (SD) and the time-dependent Temporal Domain (TD) during the training phase, along with the approximate derivative of spike activities, this method effectively resolves the convergence stability issues in SNN training. Subsequently, further studies have enabled the training of deeper SNNs \cite{zheng2021going,kim2021n} on larger-scale datasets. Regarding SNN architecture design, one study \cite{kim2022neural} proposed using Neural Architecture Search (NAS) to automatically discover optimal SNN structures. Additionally, several works have integrated attention mechanisms \cite{zhang2022wi} and Transformer architectures \cite{xing2023svformer} into SNN-based recognition networks, facilitating high-performance and stable SNN training. Therefore, these fundamental SNN techniques have been widely applied to various event-based action recognition tasks \cite{gao2023action, gao2024hypergraph, yang2024event}. However, research on MAV-specific action recognition using SNNs remains limited, as most studies focus on general event-based action recognition.


\subsection {Event-Based MAV Motion Segmentation with Time Series }
The original event stream consists of multiple MAV motions. When different MAV motions are combined and assigned corresponding semantic information, they are collectively referred to as an action. To achieve effective recognition of MAV actions, it is necessary to perform accurate segmentation of the MAV motion sequence. Event-based MAV motion segmentation involves parsing sparse event streams into discrete motion segments, each representing a specific MAV maneuver. This task presents unique challenges due to  asynchronous nature and strict real-time requirements of event data. Traditional time series analysis methods provide a computationally efficient framework suitable for this task. The foundational work \cite{velicer2003time} first established key principles for time series segmentation, though not specifically for event data. Then, a work \cite{kehagias2006time} advanced the field by applying Hidden Markov Models (HMM) to time series segmentation, demonstrating improved pattern recognition capabilities. Further developments came from a work \cite{aminikhanghahi2017survey}, which introduced change-point detection techniques for identifying statistical shifts in time series data. 

While traditional time series methods, such as HMM \cite{kehagias2006time} and change-point detection \cite{aminikhanghahi2017survey}, provide theoretical foundations, they are not optimized for the sparse and asynchronous nature of event data, necessitating specialized approaches for MAV motion segmentation. Our motion segmentation model process event streams by first converting them into event-frame representations. For each event frame, we calculate the total number of events to construct a one-dimensional temporal sequence. This time sequence enables us to effectively segment MAV motion sequence in event stream signal, achieving robust, real-time segmentation for MAV motion sequences.

\begin{figure*}[t] 
		\centering 
		\includegraphics[width=0.8\textwidth]{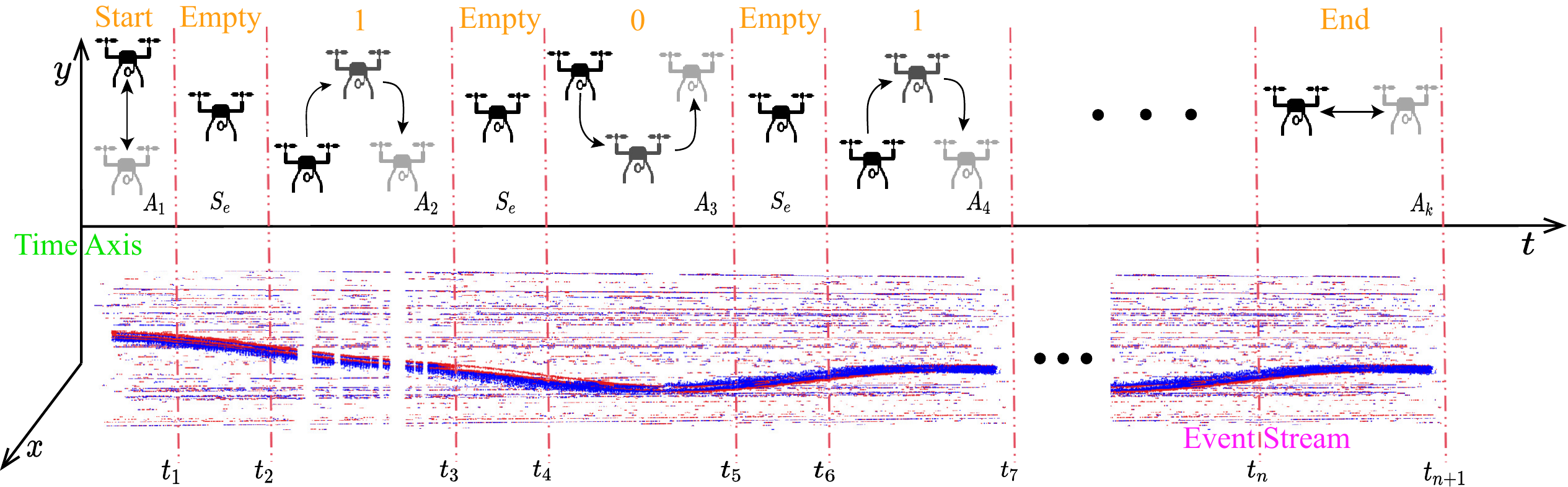}
            
		\caption{MAV motion event stream. This diagram depicts the motion flow of a complete visual message, where $A_k$ represents predefined MAV motions, $S_e$ denotes the MAV static state (Empty signal), and $t_n$ indicates the termination timestamp of each MAV motion. From a time-domain perspective, an  event signal accumulates progressively from left to right. The final ``end'' motion signifies the stop of the visual signal. Besides, the $x$-axis and $y$-axis represent the spatial domain of the event camera, while the time axis $t$ represents the time domain of the event camera.} 
        \label{fig:UAVas}
\end{figure*}

\section{The proposed method}
The proposed framework comprises three key modules. First, the MAV motion segmentation model processes the event stream captured by the event camera and segments the MAV motion sequences into individual actions. Second, the MAV action recognition model efficiently identifies each segmented MAV action and outputs the corresponding semantic label. Finally, the MAV motion sequence decoding algorithm integrates the motion segmentation and recognition models to achieve robust visual action information decoding among MAVs.

\subsection{Event-Based MAV Motion Segmentation}
To achieve visual communication based on event vision, transforming event signals into visual messages is of critical importance. However, in the communication process among MAVs, both the signal length and the transmitted information are uncertain, making it difficult to design an end-to-end model that can handle these variabilities directly.  To address this, we propose a MAV motion segmentation model that processes variable-length event streams to extract semantically meaningful motion segments, enabling reliable MAV visual communication.

\subsubsection{MAV Motion Segmentation Definition} 
MAV motion segmentation aims to detect and segment different MAV actions from a variable-length event stream $E$, determining the precise time intervals $t$ for each MAV motion. Fig. \ref{fig:UAVas} shows MAV action sequence in an event stream $E$. Then, MAV motion segmentation problem can be formulated in Eq. \ref{eq:SegDef}.

\begin{equation}
\label{eq:SegDef}
\begin{split}
S = f_{seg}(E) = \{(t_s^i, t_e^i, a^i) \mid i = 1, 2, ..., N \},
\end{split}
\end{equation}
\noindent
where $E = \{e_k=(x_k,y_k,t_k,p_k)\}_{k=1}^M$ represents the input event stream, consisting of $M$ events. Each event is defined by its spatial coordinates $(x_k,y_k)$, timestamp $t_k$, and polarity $p_k$. $S$ is the segmented motion output set, containing $N$ motion segments. In an event sequence, the start time of the $i$-th action is defined as $t_s^i$, the end time as $t_e^i$, and the action label as $a^i$.

\subsubsection{The Proposed MAV Motion Segmentation Model} 
To clearly present our event segmentation model, our event segmentation model is divided into several steps, including \textit{event count extraction}, \textit{event feature computation}, \textit{motion segmentation}, and \textit{motion refinement}.
 
\textit{Event count extraction} involves computing statistical information from the event frame sequence. In order to generate event frames, a fixed $33$ millisecond time window is used, facilitating human visual interpretation and providing intuitive results.  In segmentation method,  algorithm primarily extracts the positive event count $P_n$ negative event count $N_n$, and total event count $Q_n = P_n + N_n$ for $n$-th event frame. The frame number $frame_n$ is assigned using the line number, starting from 0. With a total frame count of $L$, this frame sequence is represented in Eq. \ref{eq:dataProcess}.
\begin{equation}
\label{eq:dataProcess}
\begin{split}
     frame_n \in \{0, 1, \dots, L-1\}, \quad n = 0, 1, \dots, L-1.
\end{split}
\end{equation}
\noindent
This step converts 2D event frames into 1D temporal signals while preserving data integrity and alignment within the event frames.

\textit{Event feature computation} effectively utilizes the extracted frame-wise event count information to analyze the MAV's motion state. To distinguish MAV motion periods from static periods, the model extracts two key features: the Positive-to-Negative Event Ratio (PNER), which measures the proportion of positive to total events, and the Event Frame Variance (EFV), which quantifies fluctuations in event counts. PNER can be defined as $R_n$ in Eq. \ref{eq:FeatureExtraction}.
\begin{equation}
\label{eq:FeatureExtraction}
\begin{split}
     R_n = \frac{P_n}{Q_n}, P_n \leq Q_n.
\end{split}
\end{equation}

Positive events $P_n$ and negative events $N_n$ typically correspond to different physical phenomena, such as changes in lighting or motion direction. $Q_n$ is the total number of event in $n$-th frame. $R_n$ highlights variations in event polarity, which may indicate specific MAV's motions, such as MAV turning or acceleration. This provides important cues for identifying MAV action start and end points in the time axis of an event steam. As for EFV, it measures the local fluctuations in total event count, reflecting the intensity or stability of MAV activity, which is key to distinguishing static and motion periods from the event stream. The specific EFV ($V_n$) is defined in Eq. \ref{eq:FeatureExtractionTotal}.

\begin{equation}
\label{eq:FeatureExtractionTotal}
\begin{split}
     V_n = \frac{1}{W} \sum_{k=n-\lfloor W/2 \rfloor}^{n+\lfloor W/2 \rfloor} (T_k - \bar{T}_n)^2, \quad \bar{T}_n = \frac{1}{W} \sum_{k=n-\lfloor W/2 \rfloor}^{n+\lfloor W/2 \rfloor} T_k,
\end{split}
\end{equation}
where $ W / 2 $ is the half-window size, and boundary frames are handled by truncation.  $W$ is a sliding window of size (default 10 frames). $T_k$ denotes the total event count at frame $k$ ($ k \in \{ n-\lfloor W/2 \rfloor, \dots, n+\lfloor W/2 \rfloor \} $), while $\bar{T}_n$ is the mean value of total event number in a window $W$. Besides, $V_n$ measures the rate of event count variation, with lower values in static periods and higher values in MAV motion periods. To reduce noise, $R_n$ and $V_n$ are smoothed using a 5-frame moving average, which enhances the model's ability to detect MAV motion boundaries.

\textit{Motion Segmentation} step identifies MAV motion segments by applying threshold-based segmentation to the smoothed $R_n$ and $V_n$. It detects transitions between static and motion periods and generates initial MAV's motion boundaries based on predefined criteria, such as minimum motion duration. To segment MAV motions, event frames in the static state are first identified. In the implementation process, Eq. \ref{eq:labelFrame} is applied to label the state of each event frame.

\begin{equation}
\label{eq:labelFrame}
\begin{split}
G_n = 
\begin{cases} 
1, & \text{if } R_n < \theta_R \text{ and } V_n < \theta_V, \\
0, & \text{otherwise}.
\end{cases}
\end{split}
\end{equation}

In Eq.~\ref{eq:labelFrame}, $\theta_R$ is equal to 0.5 and $\theta_V$ is set to $50\%$ of the median variance $V_n$, where $G_n = 1$ is static frame and $G_n = 0$ represents motion frame. To search for MAV motion boundaries, boundary detection computation is conducted.
\begin{equation}
\label{eq:boundary}
\begin{split}
D_n = G_{n+1} - G_n, \quad B = \{ b_j \mid D_n \neq 0 \},
\end{split}
\end{equation}
where $D_n$ is a differential sequence, and $B$ is the boundary point set. Using Eq. \ref{eq:boundary}, a sequence of boundary pairs in the event stream is obtained. Then, the $G_{b_j} = 0 \land b_{j+1} - b_j \geq L_{min} , L_{min} = 10$ is applied to filter out motion segments shorter than 10 frames (approximately 330 milliseconds), as every motion in our MAV communication exceeds this threshold. Finally, these motion segments are recorded as candidate motion segments.

\textit{Motion refinement} enhances the detection of precise MAV motion moments. In this step, short-duration MAV motions are filtered out by retaining only segments with a duration of at least 30 frames (approximately 1 second, given a frame rate of 33 ms), ensuring the removal of transient noise. Then, adjacent motion segments separated by gaps of 10 frames or fewer (approximately 330 ms) are merged into a single continuous segment, enhancing temporal coherence. Finally, MAV segments shorter than 91 frames (approximately 3 seconds) are discarded to ensure that only motions of sufficient duration are preserved, aligning with the expected characteristics of meaningful MAV activities, e.g., action ``1'', action ``0'', action ``start'', and action ``end''.

\begin{figure*}[t] 
		\centering 
		\includegraphics[width=0.8\textwidth]{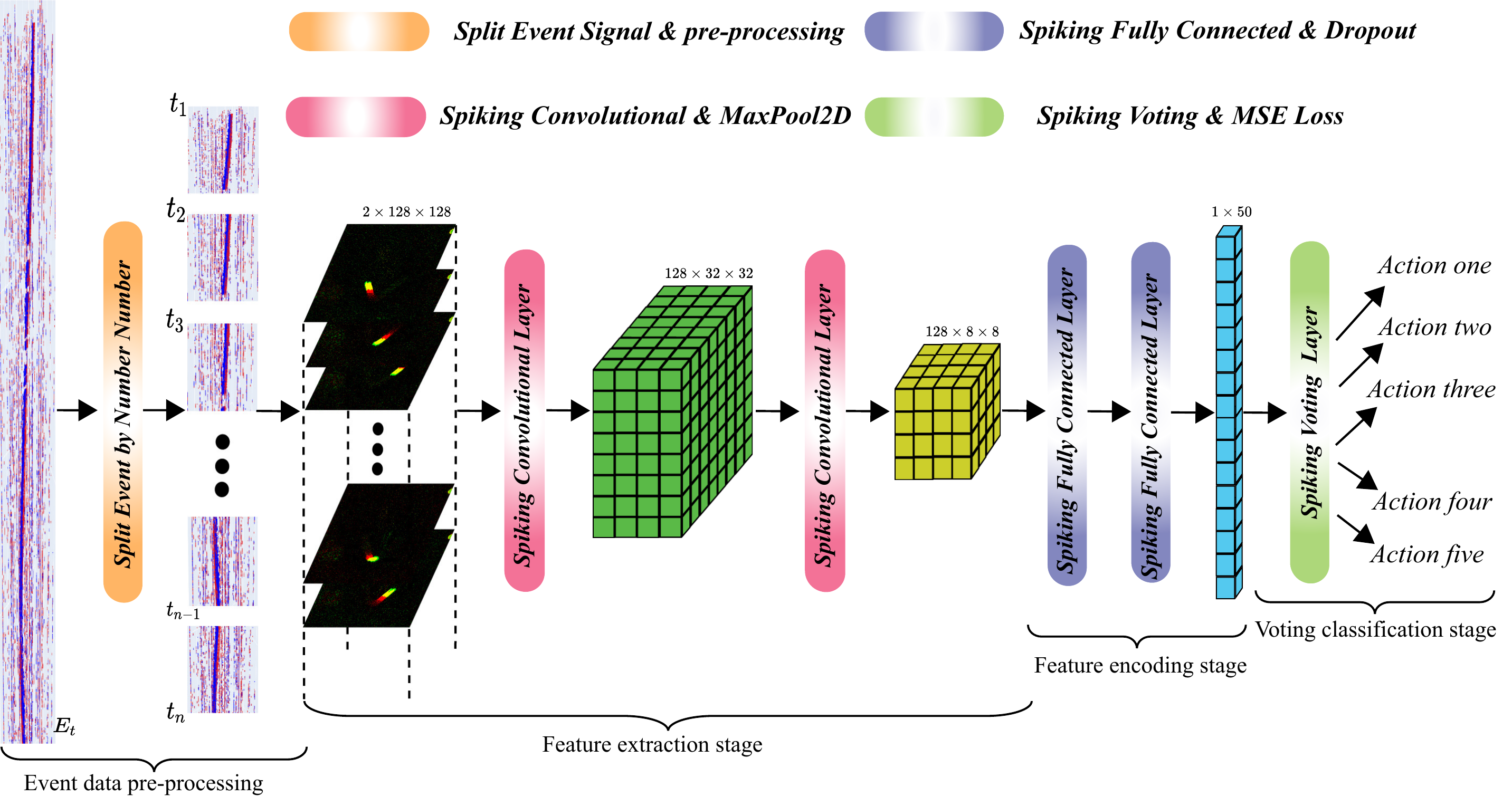}
            
		\caption{EventMAVNet: MAV action recognition model. This diagram illustrates the complete recognition process of the MAV event stream, implemented as a simple yet effective five-layer spiking neural network, encompassing four main components: event data pre-processing, feature extraction stage, feature encoding stage, and voting classification stage. } 
        \label{fig:recognitionModel}
\end{figure*}


\subsection{MAV Action Recognition Network}
To recognize each segmented MAV action from the event stream collected by an event camera, a compact and efficient spiking neural network was developed to enable rapid and accurate MAV action recognition.

\subsubsection{The basic theories of Spiking Neural Network}
Spiking Neural Networks (SNNs) \cite{tavanaei2019deep} are a class of artificial neural networks 
inspired by the behavior of biological neurons, where information is transmitted via discrete spikes or events rather than continuous values. Using the classic Leak Integrate-and-Fire (LIF) neuron \cite{stoliar2017leaky} model as an example, a neuron integrates incoming signals and emits a spike ($S(t) \in \{0,1\}$) when its membrane potential $H(t)$ exceeds a threshold $H_{th}$. The LIF neuron's behavior is describe in Eq. \ref{eq:snn}.

\begin{equation}
\label{eq:snn}
\begin{split}
  & H(t+1) = H(t) \cdot e^{-\frac{\Delta t}{\tau_m}} + R \cdot I(t) \\
 &  I(t) = \sum_j{w_j \cdot S_j(t)}  \\
 & S(t) =
\begin{cases} 
1, & \text{if } H(t) \geq H_{th}, \\
0, & \text{otherwise}.
\end{cases}
\end{split}
\end{equation}
In Eq. \ref{eq:snn}, the membrane potential $H(t)$ decays over time with a time constant $\tau_m$ and increase with the input current $I(t)$, scaled by resistance $R$. The input $I(t)$ is the weighted sum of spikes from pre-synaptic neurons, with $w_j$ as synaptic weights. A spike $S(t)=1$ occurs if $H(t)$ reaches $H_{th}$, otherwise $S(t)=0$. This step captures the event-driven essence of SNNs.

\subsubsection{Spiking Neural Networks for Event-Based Processing and MAV Action Recognition Model}
Traditional Deep Neural Networks (DNNs) recognize event motion stream in a frame-based, synchronous manner, relying on continuous activation values to capture spatial patterns. Traditional DNNs, while effective for frame-based inputs, struggle with the asynchronous, sparse nature of event streams, leading to high computational costs and inefficiency in real-time MAV action recognition. For real-time applications, such as MAV action recognition, fast, efficient, and low-power pattern discrimination becomes particularly crucial. In contrast, SNNs offer a biologically inspired approach by representing information through discrete spike trains (temporal sequences of binary events). Compared to DNNs, recognition models based on SNN possess several advantages. First, unlike DNN neurons that output continuous values \cite{ying2021psigmoid}, SNN neurons emit spikes only when the membrane potential exceeds a threshold. Thus, during MAV action recognition, only a small fraction of neurons are activated, granting SNNs a significant advantage in power efficiency. Second, the spiking mechanism of SNNs inherently integrates input signals over time, enabling them to capture temporal dynamics in event-based data. This makes SNNs particularly suitable for processing time-varying signals, such as those event streams from event cameras. Although our method converts event data into frames for structured input, SNNs outperform traditional DNNs in power efficiency, temporal dependency modeling, and parameter efficiency. 


Based on these properties, EventMAVNet, a spiking neural network tailored for MAV action recognition, is proposed. The EventMAVNet architecture processes event streams with an input resolution of 128$\times$128 and $2$ channels (representing positive and negative polarity events) over 16 time steps, classifying them into 5 distinct motion categories. As illustrated Fig. \ref{fig:recognitionModel}, the proposed networks is structured into three stages: \textit{feature extraction}, \textit{feature encoding}, and \textit{voting classification}, all implemented using the SpikeJelly framework \footnote{https://github.com/fangwei123456/spikingjelly}.

Feature extraction stage extracts spatial-temporal features from the preprocessed event frames. It consists of two spiking convolutional layers, each with 3$\times$3 kernel and 128 outputs channels. The first layer processes the 2-channel input of size 128$\times$128, while the second builds upon the previous output, maintaining 128 channels. Each convolutional layer is followed by a Batch Normalization (BN) layer to stabilize the distribution of spiking activations across the 16 time steps, enhancing training convergence. Leaky Integrate-and-Fire (LIF) neurons then introduce temporal dynamics, integrating a 4$\times$4 max-pooling operation downsamples the feature maps, reducing the spatial resolution to 32$\times$32 after the first layer and 8$\times$8 after the second. This results in a feature tensor of shape 128$\times$8$\times$8 per time step.   

Feature encoding stage transforms the extracted features into a compact, discriminative representation. The feature maps are flattened into a 128$\times$64-dimensional vector (computed as 128 channels $\times$8$\times$8 spatial grid), followed by a dropout layer ($p=0.5$) to prevent overfitting. A spiking fully connected layer reduces this vector to 128 dimensions, with LIF neurons processing the temporal spikes across 16 time steps. Another dropout layer further regularizes the output, ensuring robustness in the encoded features. 
 
Voting classification stage maps the encoded features to action classes and computes the classification output. A spiking fully connected layer projects the 128-dimensional features into 50 dimensions. A spiking voting layer then aggregates the spiking activity of these 50 neurons, organized as 10 neurons per class to produce outputs for 5 action categories. The five-class classifier (four semantic labels + background noise) improves decoding robustness by enabling noise filtering during recognition, ensuring cleaner MAV motion sequence decoding.

Finally, the Mean Squared Error (MSE) loss is then calculated to measure the difference between this averaged prediction and the target labels across a batch in training.
\begin{equation}
\label{eq:loss}
\begin{split}
L = \frac{1}{O} \sum_{o=1}^{O} \sum_{p=1}^{P} (\text{pred}_{o,p} - y^{l}_{o,p})^2.
\end{split}
\end{equation}
In Eq. \ref{eq:loss}, $O$ is the batch size, $P$ is the number of classes, $\text{pred}_{o,p}$ is the predicted score for the $o$-th sample and $p$-th class after temporal averaging, and $y^{l}_{o,p}$ is the corresponding one-hot value (number $0$ or $1$). 

\subsection{IMSR: Integrated Segmentation and Recognition for MAV Motion Decoding}
To improve the reliability of MAV visual communication, we propose an online decoding framework termed \textit{Integrated MAV Segmentation 
and Recognition (IMSR)}. As shown in Algorithm \ref{decodingFramework}, IMSR decodes MAV-transmitted messages from raw event streams captured by an event camera. The framework operates in three main stages: preprocessing, segmentation, and recognition. In the preprocessing stage, the background noise filter $\mathcal{F}$ removes spurious events caused by environmental noise or camera artifacts, using a threshold-based approach to retain valid motion-related events.
The denoised event stream $D^e_f$ is buffered for subsequent processing. The segmentation stage begins by monitoring the number of valid event frames. Then, valid event frames are counted by \texttt{countEventFrames()} in the filtered stream. Once the accumulated count exceeds a threshold (e.g., 100), the segmentation model $Seg$ is activated to partition the event stream into discrete motion segments $Action_k$, where each segment is assumed 
to correspond to a distinct MAV action. The threshold is set based on the minimum duration of predefined MAV actions (approximately 3 seconds). Given that the event stream is sampled at a rate of 30 frames per second, the threshold of 100 ensures sufficient temporal context (approximately 3.3 seconds) to capture at least one complete action, thereby reducing the risk of segmenting incomplete or noisy motion sequences.

In the recognition stage, each segmented action $a_i \in Action_k$ is classified by the recognition model $Rec$, implemented as EventMAVNet. Recognized labels $r_i$ are appended sequentially to a code sequence $C_{seq}$, when $r_i$ is not equal to label ``background''. If the keyword ``end''  is detected as the final label, decoding is triggered. The \texttt{decode()} function extracts a three-part message $M_{dec}$ from $C_{seq}$, including flight direction, encoded angle, and relative distance. For message integrity, decoding only proceeds if the sequence starts with ``start'' and reaches the minimum expected length. By combining temporal segmentation with learned action recognition, IMSR ensures robust decoding even under noisy or ambiguous visual input. Unreliable inputs that fail segmentation or recognition are discarded, and retransmission is requested, enabling reliable MAV visual communication in dynamic environments.     

\begin{algorithm}[t] 
\caption{MAV motion sequence decoding algorithm} 
\label{decodingFramework} 
\begin{algorithmic}[1]
        \STATE {\textbf{Input:}} The event stream \(D^e\) collected by event camera
        \STATE {\textbf{Output:}} Communication code $C_{seq}$, decoded message $M_{dec}$ 
        \STATE Initialize background noise filter $\mathcal{F}$
        \STATE $\mathcal{F}.\text{accept}(D^e)$
        \STATE $D^e_f \gets \mathcal{F}.\text{generateEvents}()$
        \STATE Buffer the event stream $D^e$ in memory
        \STATE Set $C_{seq} \gets [\,]$, $M_{dec} \gets [\,]$
        \STATE Initialize segmentation model $Seg$, recognition model $Rec$
        \STATE Initialize segmented action set $Action_k$
        
        \WHILE{new frame features available from $D^e_f$}
            \STATE $E_c \gets \text{countEventFrames}(D^e_f)$
            \IF{$E_c > 100$}
                \STATE $Action_k \gets Seg.segment(D^e_f)$  
                \FOR{\textbf{each} $a_i$ \textbf{in} $Action_k$}
                    \STATE $r_i \gets Rec.predict(a_i)$  
                    \IF{$r_i$ $!= \texttt{"background"}$}
                        \STATE Append $r_i$ to $C_{seq}$
                    \ENDIF
                \ENDFOR
                \IF{$C_{seq}[-1] == \texttt{"end"}$}
                    \STATE $M_{dec} \gets decode(C_{seq})$
                    \STATE \textbf{break}
                \ENDIF
            \ENDIF
        \ENDWHILE
    \STATE $M_{dec} \gets$ decode($C_{seq}$)
    \RETURN $C_{seq}, M_{dec}$

    \vspace{0.5em}
\STATE \textbf{Function} $decode(C_{seq})$
\IF{$length(C_{seq}) < 7$ \textbf{or} $C_{seq}[0] \neq \texttt{"start"}$}
    \STATE \textbf{return} $[\,]$
\ENDIF
\STATE $direction \gets C_{seq}[1]$
\STATE $angle \gets C_{seq}[2:5]$
\STATE $distance \gets C_{seq}[5:7]$
\STATE $M_{dec} \gets [direction, angle, distance]$
\STATE \textbf{return} $M_{dec}$
\end{algorithmic} 
\end{algorithm}

\section{Experiments}
This section provides the evaluation of our proposed model. We first outline the experimental details, which include information about the dataset, parameter settings, and experimental environment\footnote{Dataset and code will be made publicly available upon publication.}. Then, our comparative experiments assess both MAV action recognition (accuracy and speed) and motion segmentation performance separately. Additionally, we conducted key ablation experiments on the proposed action recognition model to offer a detailed analysis of the contributions of its critical components. Finally, we describe several experiments on the visual communication of common information to validate the effectiveness of our entire MAV-based visual communication system. 

\begin{figure}[t] 
		\centering 
		\includegraphics[width=0.48\textwidth]{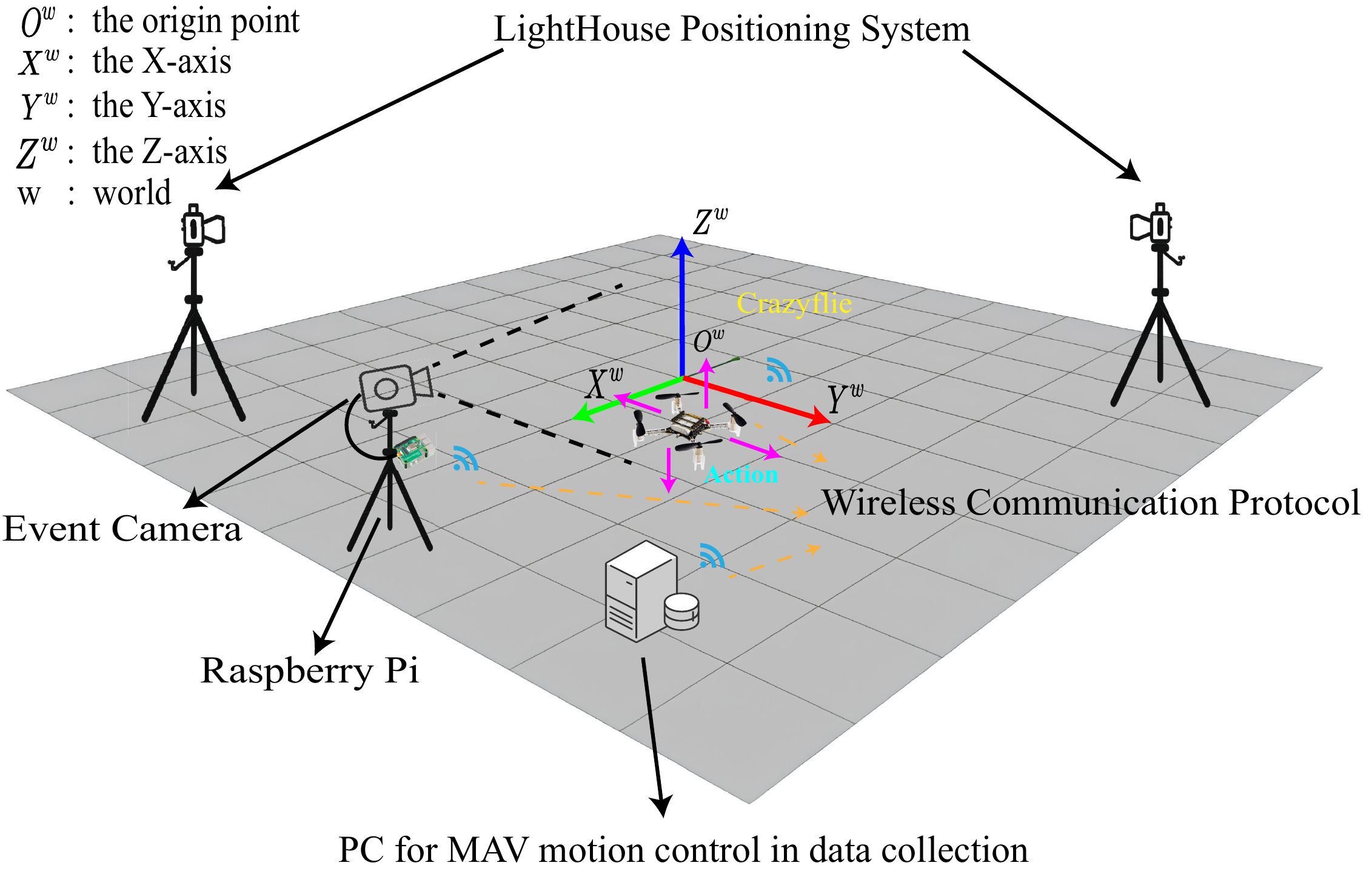}
		\caption{MAV motion data collection system. The figure lists the necessary data collection sensors and key processing devices.} \label{dataCollect}
\end{figure}

\subsection{Experimental details}
To validate the proposed framework, a controlled MAV motion data collection system in an indoor environment is designed to evaluate visual communication behavior and motion segmentation performance between MAVs, as shown in Fig.~\ref{dataCollect}. In the figure, the hardware consists of Crazyflie drones in the center of the experimental scene, a DVS Micro Explore event camera on a tripod to capture event data, a Lighthouse system overhead for precise 3D localization, and a PC with an NVIDIA GPU 4090 running Ubuntu for data processing. The DVS camera recorded event streams in aedat4 format, which are then processed using the DVS process library. During the collecting process, the data is collected at three sensor observation distances (0.9~m, 1.2~m, and 1.5~m), representing short, medium, and long ranges, respectively, using an ``Observe'' and ``Motion'' feedback loop, in which the camera captured the MAV's motions while the PC processed the data for MAV control and analysis. 

The dataset consists of event streams capturing MAV motions at these three scales, with each scale including five action categories: Action ``0'', Action ``1'', Action ``Start'', Action ``End'', and Action ``background''. Particularly, the MAV motion background signal helps the model distinguish between normal signals and non-effective signals (included empty and noise signals), enabling robust action classification.       

As for the MAV motion segmentation model, we implemented our method using Python 3.10 with the NumPy library. The model's performance was evaluated using center point error (the temporal deviation of predicted action midpoints) and 2D Intersection over Union (IoU) error (the overlap ratio between predicted and ground-truth action segments) to ensure a fair evaluation of the MAV motion segmentation model's performance. With regard to the MAV action recognition model, we built the framework using PyTorch and the SpikeJelly library, leveraging their capabilities for processing event-based data. The recognition model is evaluated using classification accuracy, model parameter size, inference speed, and power consumption, ensuring a fair and impartial assessment that guides the design of a more efficient and robust MAV visual communication system.

\subsection {MAV Action Recognition and Inference Speed Experiment}

To evaluate the effectiveness of the proposed MAV action recognition approach, we conducted a MAV action recognition experiment using event data with other state-of-the-art model ( spike-C \cite{eshraghian2023training}, DVS-G \cite{amir2017low}, spike-R \cite{fang2021deep}). The test results are presented in Table \ref{tab:actionRec}.

\begin{table}[h]
    \scriptsize
    \centering
    \caption{Performance Comparison Across Different Observation Distance Datasets.}
    \label{tab:actionRec}
    \begin{tabular}{cccc}
    \hline
    Algorithms & Short & Medium & Long \\ \hline
    spike-C \cite{eshraghian2023training}   &$94.63\%\pm 1.82\%$ &$93.00\%\pm 1.64\%$  & $91.49\%\pm 8.2\%$  \\ 
    spike-R \cite{fang2021deep} &$96.49\%\pm 0.47\%$&$94.54\%\pm 0.24\%$&$92.35\%\pm 1.23\%$\\
    DVS-G \cite{amir2017low}   &$95.36\%\pm 1.4\%$&$94.12\%\pm 0.6\%$&$91.82\%\pm 1.56\%$    \\
    Ours  & $96.51\%\pm 0.61\%$ &$95.37\%\pm 1.01\%$ &$94.98\%\pm 1.17\%$      \\
    \hline
    \end{tabular}
\end{table}

\begin{figure}[t]
		\centering 
		\includegraphics[width=0.38\textwidth]{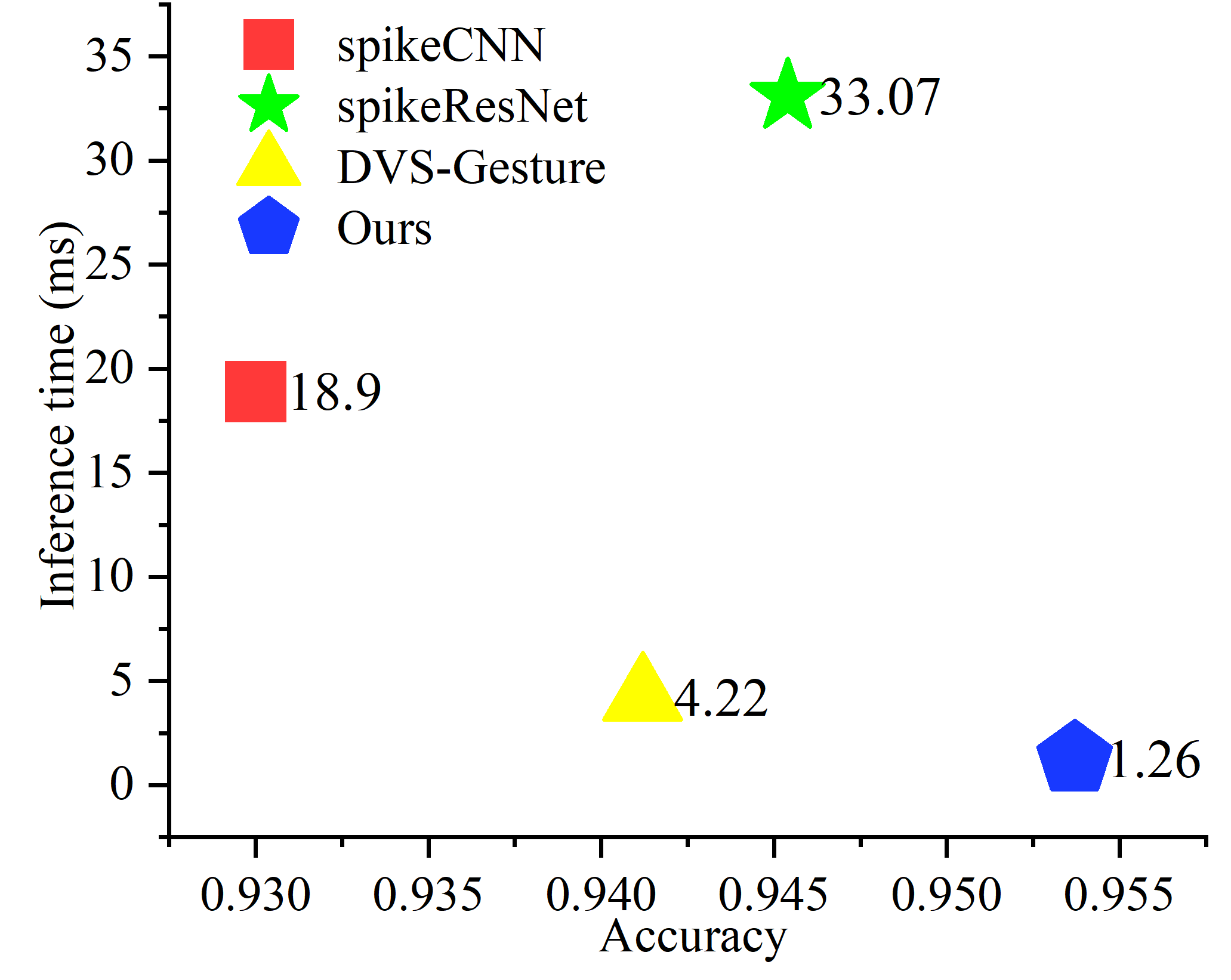}
		\caption{Inference time comparison with a batch size of 10. Each result represents the average runtime over 100 runs on an RTX 4090 GPU.} \label{fig:inferenceTime}
\end{figure}

As demonstrated in Table \ref{tab:actionRec}, our model outperforms other state-of-the-art methods across the short, medium, and long scenarios in terms of recognition accuracy, underscoring its advanced capabilities to a significant extent. Notably, all models exhibit a decline in accuracy in the long-distance scenario, likely due to reduced event density and increased noise at greater observation distances. Nevertheless, our model maintains a competitive recognition performance on the long dataset (1.5~m observing distance). Besides, we also present the accuracy and inference time of MAV action recognition on the Medium dataset in Fig. \ref{fig:inferenceTime}. Compared to other models, ours model achieves the lowest latency of 1.26 ms per 10 test samples, demonstrating its reliability and efficiency under demanding conditions.

\begin{figure}[h]
\centering 
\subfigure[Training and Testing Accuracy.]{%
\resizebox*{4.2cm}{!}{\includegraphics{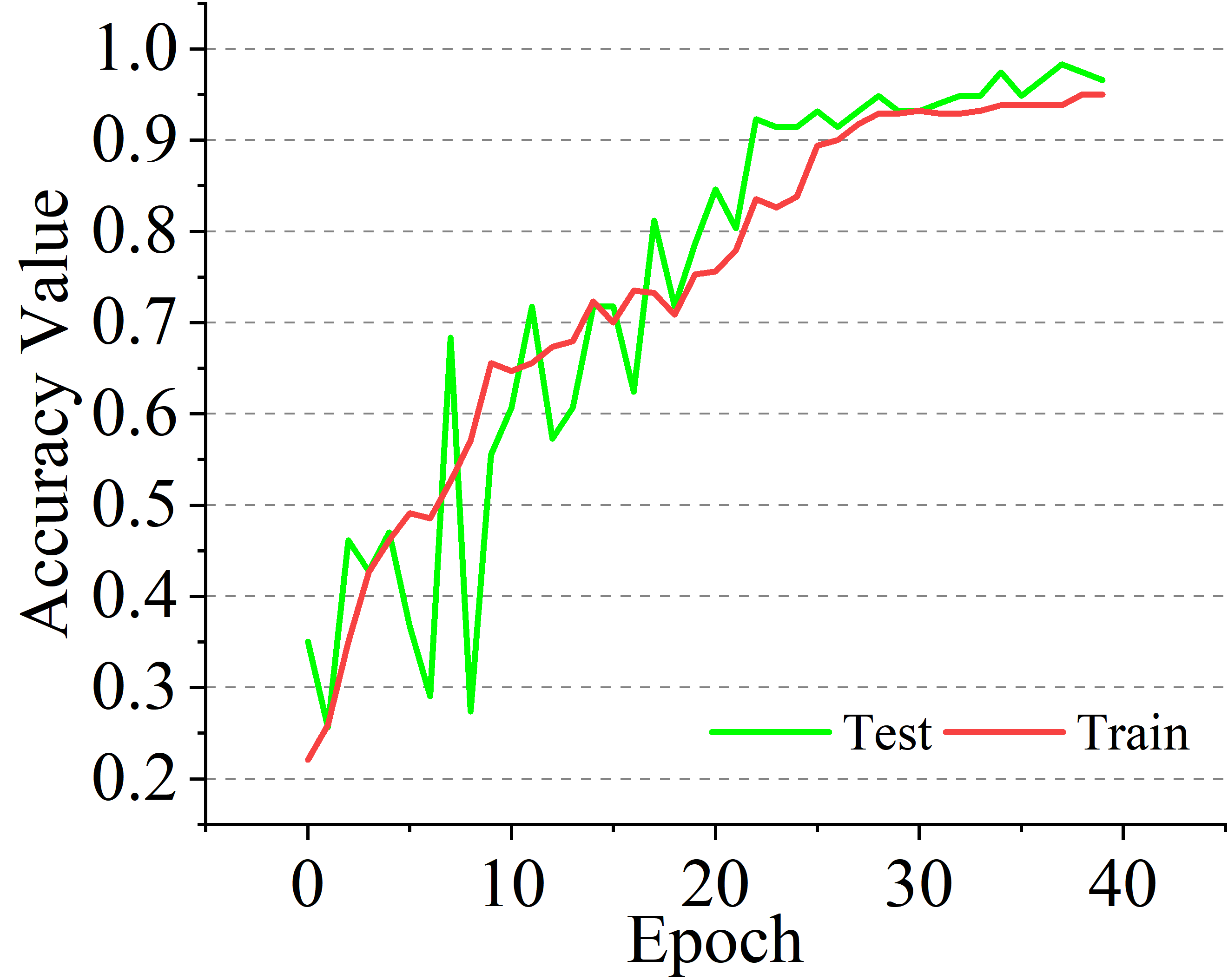}}
\label{fig:two:a}
}
\subfigure[Training and Testing Loss.]{%
\resizebox*{4.2cm}{!}{\includegraphics{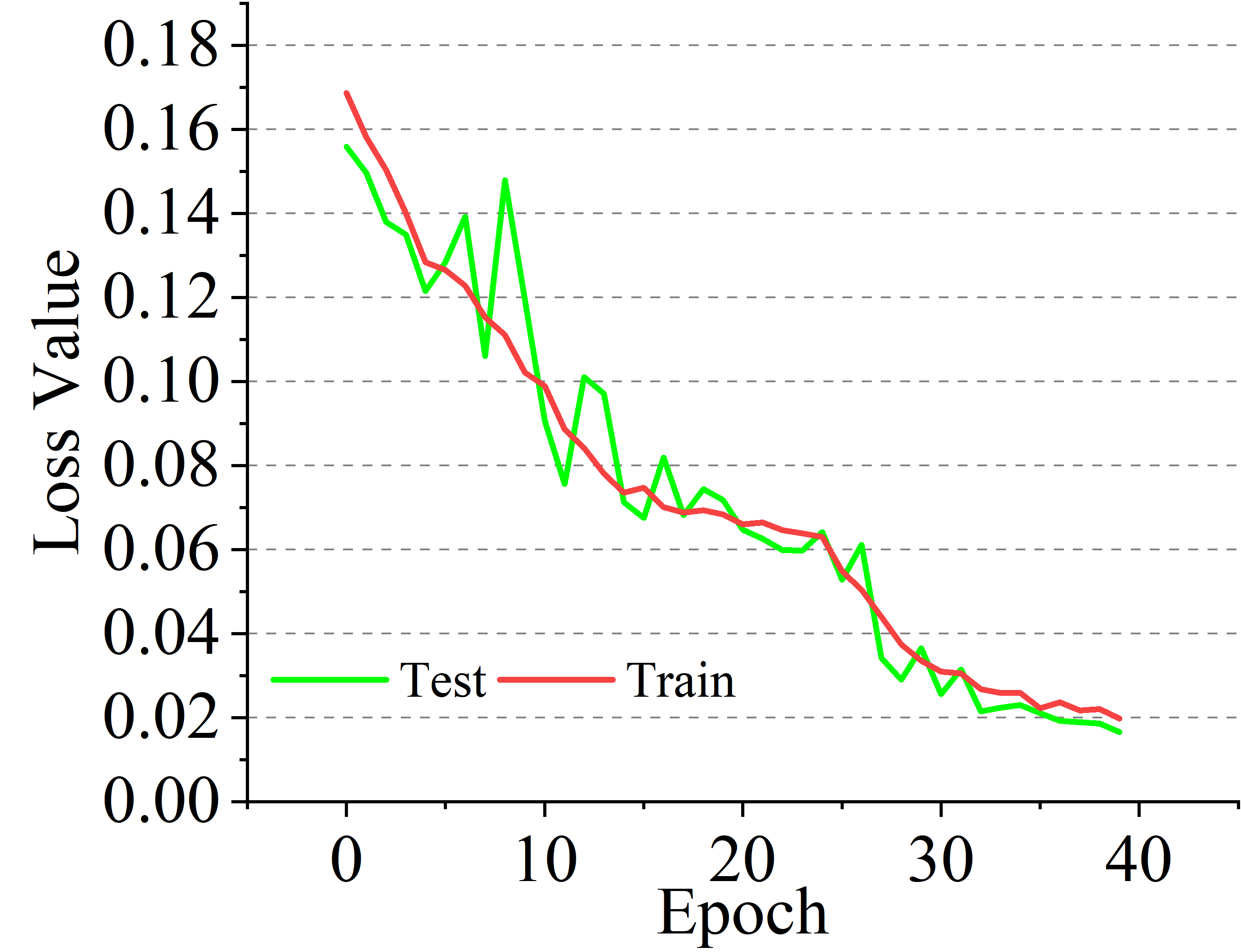}}
\label{fig:two:b}
}
\caption{Training and Testing Performance Over Epochs. (a) Training (green) and testing (red) accuracy over 40 epochs. (b) Training (green) and testing (red) loss over 40 epochs.} \label{fig:two}
\end{figure}

To evaluate the effectiveness of our proposed model for recognizing MAV action data in event form, we conduct experiments in dataset with a 3:1 training-to-testing ratio. The primary motivation is to assess the model’s tailored design, which is intended to enhance temporal precision and robustness in processing event-based MAV motion sequences for real-time applications. To this end, we visualized the training and testing performance over epochs to demonstrate the model's convergence and generalization capabilities on MAV action recognition task. Specifically, Fig \ref{fig:two} illustrates the proposed model's training and testing performance in detail and Fig \ref{fig:two:a} shows the accuracy change over 40 epochs, highlighting the model’s learning progression and generalization capability on event-based MAV action recognition data. In this figure, training is green curves, and testing is red curves. Fig. \ref{fig:two:b} shows the loss values over 40 epochs, demonstrating stable convergence and reduced overfitting. The training (green) and testing (red) curves confirm the model's robustness for temporal event sequences.

Overall, these experimental results validate the efficacy of our proposed model, as the visualizations of accuracy and loss curves demonstrate robust convergence and enhanced generalization, supporting the model's capability to accurately identify MAV in-flight actions. These findings not only confirm the potential of SNN models for event-based action recognition tasks but also pave the way for future optimizations and broader applications in real-world MAV scenarios.

\begin{figure}
		\centering 
		\includegraphics[width=0.4\textwidth]{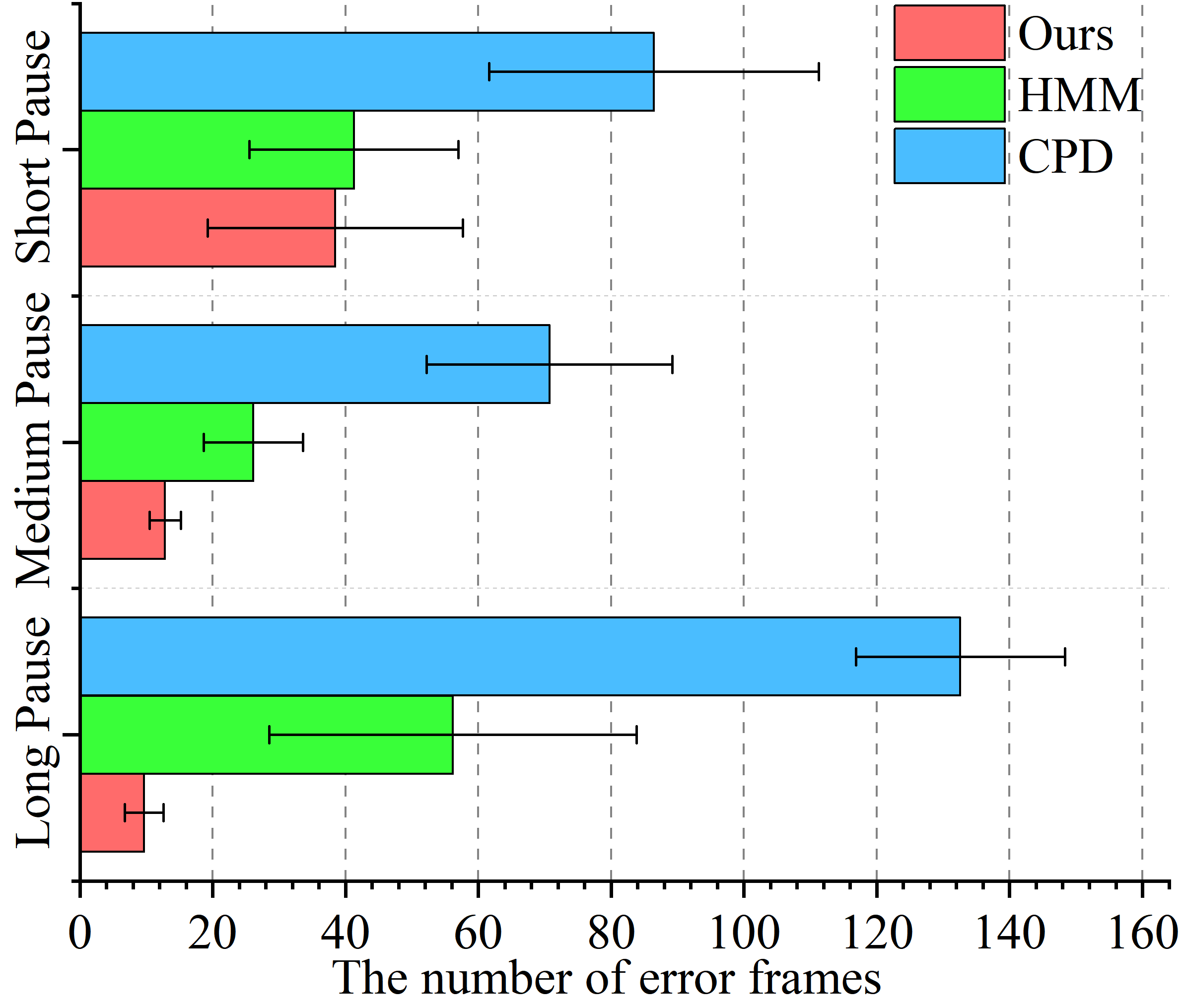}
		\caption{Comparison of the number of error frames for three methods (Ours, HMM \cite{kehagias2006time}, CPD \cite{aminikhanghahi2017survey}) under different empty signal types: short pause, medium pause, and long pause. The error reflects the misalignment in center point prediction of the MAV action.} \label{figCentreError}
\end{figure}

\subsection{The Comparison of Motion Segmentation Model Results}
A robust motion segmentation model is pivotal for ensuring reliable visual communication in MAVs. Inaccurate motion localization by the segmentation model can directly impair the recognition accuracy of subsequent MAV action classification processes. To provide a more objective assessment of the motion segmentation model proposed in this study, a comprehensive, multi-dimensional experimental comparison is conducted in following. As illustrated in Fig. \ref{fig:UAVas}, an empty signal is inserted between consecutive valid MAV actions during motion message delivery to mitigate overlap and confusion between different actions. In our comparative experiments, we evaluated three key metrics: the offset of the MAV action center point, the total frame number error of the action, and the action overlap area ratio, with HMM \cite{kehagias2006time} and CPD \cite{aminikhanghahi2017survey}. Furthermore, the model's segmentation performance under varying durations of empty signals was analyzed, providing insights into its robustness and adaptability. The experiments are conducted on three event streams, each containing the same 9 valid MAV actions and 8 empty signals. The durations of the empty signals in three streams are 2.5 seconds, 3 seconds, and 3.5 seconds, respectively. 

To evaluate the offset error of MAV action centers, we computed the center deviation $\{A_k\}_c$ and its variance $\{A_k\}_c^{var}$ for each action. Lower values indicate higher prediction accuracy. Fig. \ref{figCentreError} shows three event streams with different empty signal durations on the horizontal axis and the frame offset of the predicted center on the vertical axis. The results demonstrate that the proposed MAV motion segmentation model achieves more accurate motion localization on all three streams, with both the deviation and variance notably smaller than those of other models. 

\begin{figure}
		\centering 
		\includegraphics[width=0.4\textwidth]{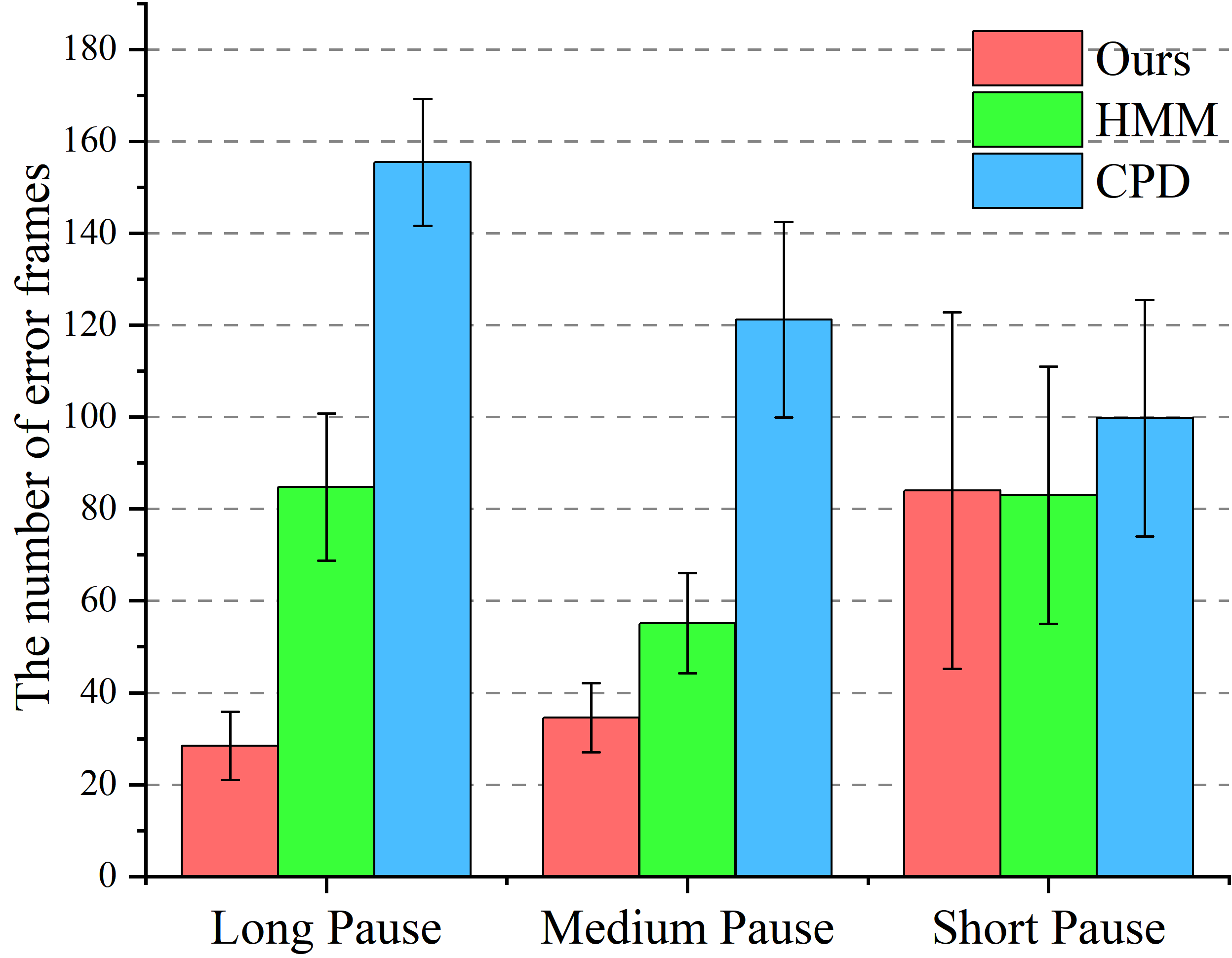}
		\caption{Comparison of total action duration errors for three methods under different empty signal types. The error represents the frame-level difference between predicted and ground-truth MAV action intervals.} \label{figTotalError}
\end{figure}

With regard to the total frame number error of the MAV action, we calculated the difference between the predicted duration  $A_k^{pred}$ and the ground truth duration $A_k^{GT}$ for each action. This metric reflects how well different models can localize the temporal span of MAV actions. As shown in Fig. \ref{figTotalError}, although our segmentation model performs slightly worse than the HMM model under the 2.5-second idle signal condition, it achieves clear advantages when the empty signal duration increases to 3 and 3.5 seconds. This indicates that our model becomes more effective at action duration prediction as the length of empty signals increases.     

\begin{figure}
		\centering 
		\includegraphics[width=0.4\textwidth]{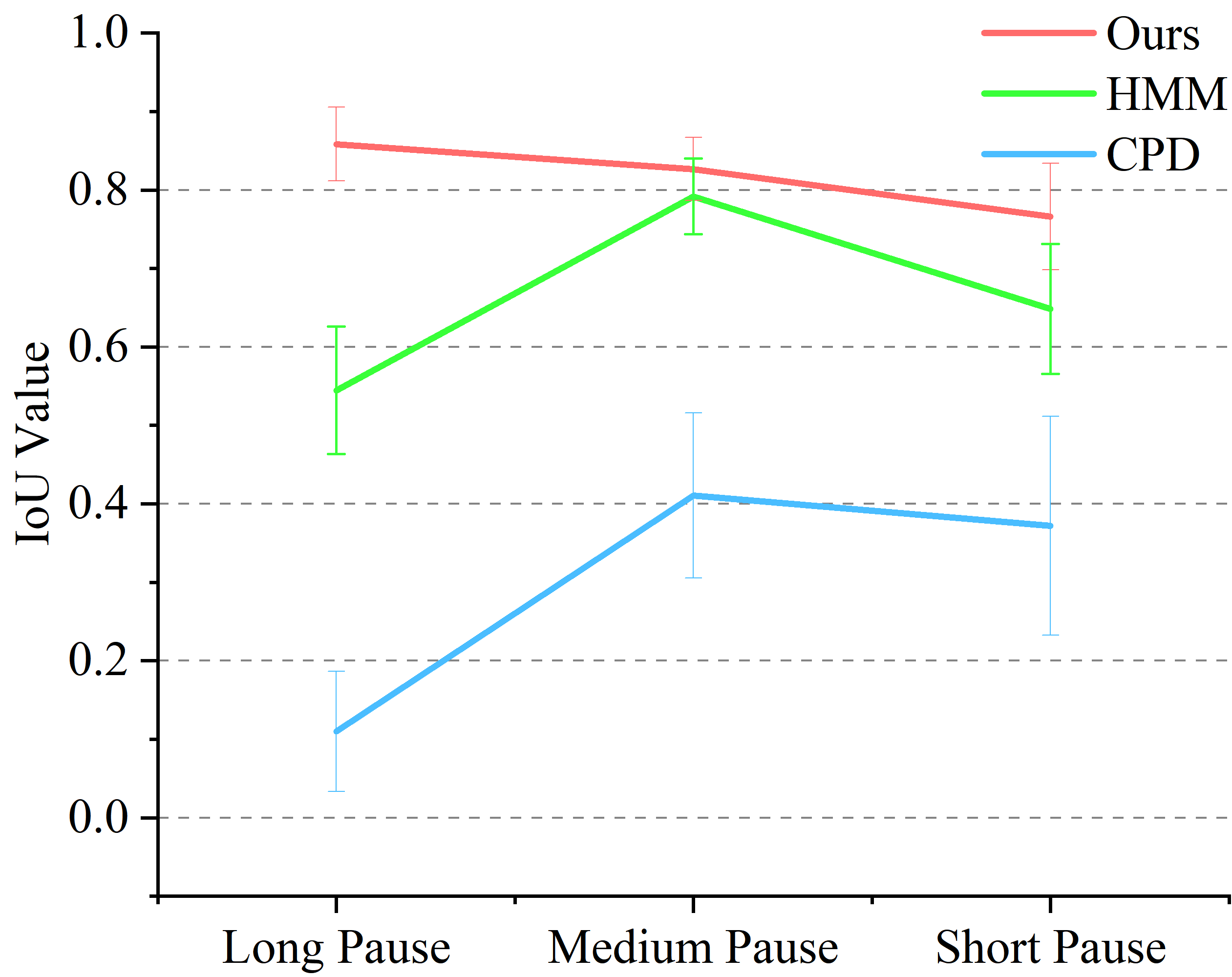}
		\caption{Comparison of intersection-over-union (IoU) errors between predicted and ground-truth MAV actions under different MAV action pause intervals.} \label{IOUError}
\end{figure}

Regarding the MAV action overlap area ratio metric, we analyze the start and end position of MAV actions to compute intersection $\cap$ and union $\cup$ of these actions. The calculation is expressed in Eq. \ref{eq:iou}, where $A^{pred}$ represents the motion segment predicted by the model, and $A^{gt}$ denotes the ground truth motion segment. The experimental results are presented in Fig. \ref{IOUError}. In the figure, the IOU test results of our proposed model consistently outperform others, demonstrating the superiority of the model proposed in this study.

\begin{equation}
\label{eq:iou}
\begin{split}
& A^{pred} = [start_{pred},end_{pred}],  \\
& A^{gt} = [start_{gt},end_{gt}],  \\
& \text{IOU} = \frac{|A^{pred} \cap A^{gt}|}{|A^{pred} \cup A^{gt}|}.
\end{split}
\end{equation}

\begin{figure}
		\centering 
		\subfigure[MAV action sequence with 2.5s pause signals.]{%
			\resizebox*{0.45\textwidth}{!}{\includegraphics{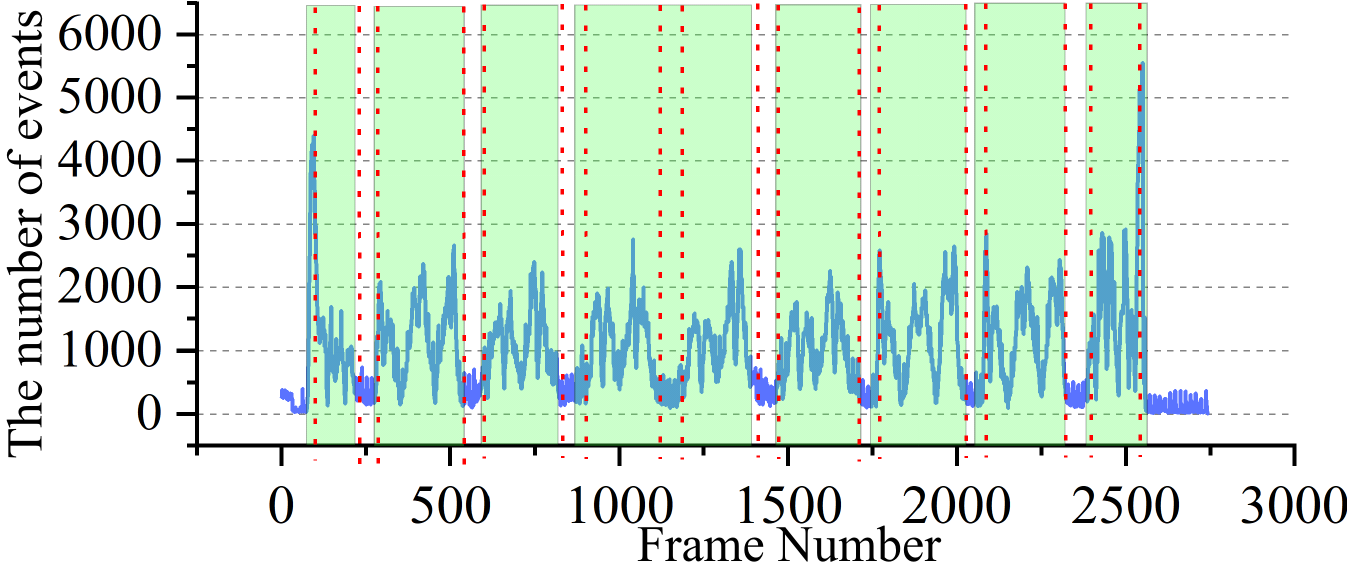}}}
		\subfigure[MAV action sequence with 3.0s pause signals.]{%
			\resizebox*{0.45\textwidth}{!}{\includegraphics{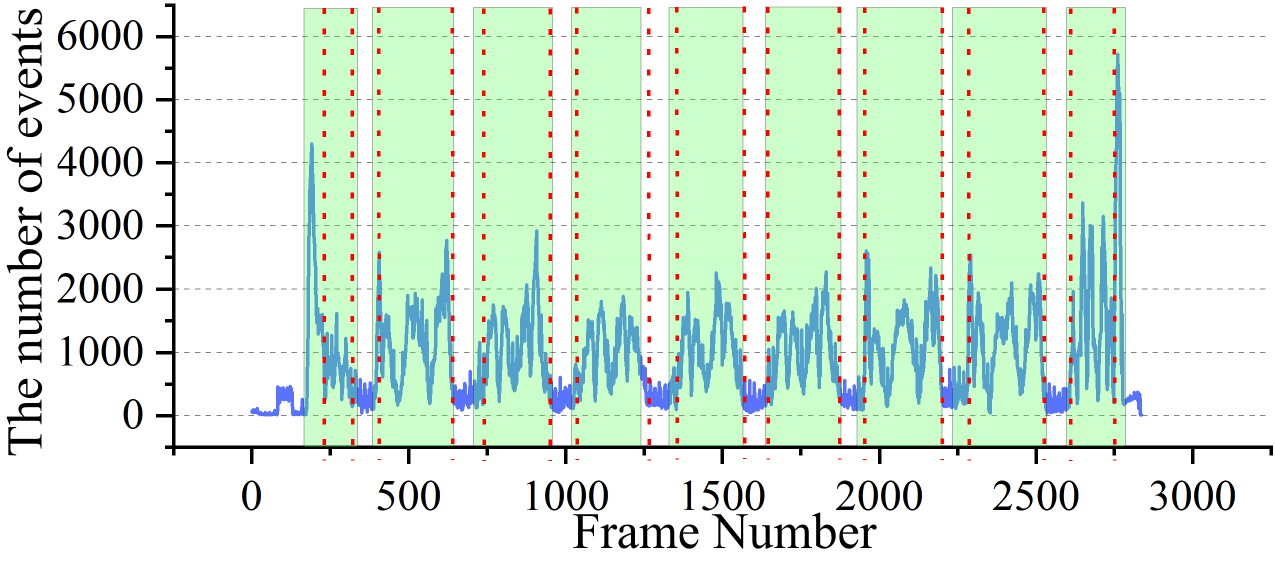}}}
            \subfigure[MAV action sequence with 3.5s pause signals.]{%
			\resizebox*{0.45\textwidth}{!}{\includegraphics{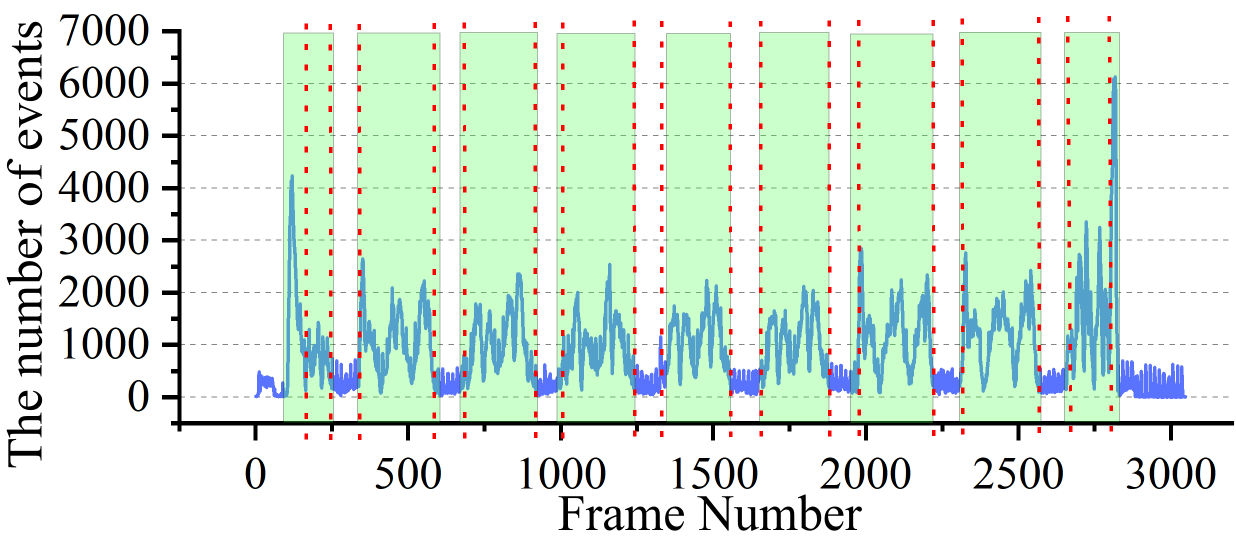}}}            
		\caption{Comparison of segmentation performance under different MAV pause durations. The red dashed lines denote the ground truth boundaries of each action, and the translucent green boxes indicate the predicted start and end positions by the proposed segmentation model.} 
  \label{fig:UAVsegSample}
\end{figure}

To intuitively demonstrate the robustness of the proposed motion segmentation model, we visualize the MAV action segmentation results on MAV action sequences under three different durations of the MAV pause signal, as shown in Fig \ref{fig:UAVsegSample}. When the pause signal duration is 2.5~s, the model incorrectly segments $9$ valid MAV actions into $8$. This error likely stems from insufficient temporal separation between actions, causing the algorithm to fail in detecting clear MAV motion boundaries due to weak or ambiguous temporal features. However, within 3~s and 3.5~s pause signal, the proposed model accurately detects the start and end of each MAV action. 


\subsection {Ablation Study on The Proposed Recognition Model}
To gain a deeper understanding of the proposed MAV action recognition model, an ablation study is conducted in this section. Specifically, we analyzed the effects of spike window length $F$ and event frame resolution on our spiking-based MAV motion recognition model. The detailed experimental results are shown in table \ref{tab:R} and \ref{tab:T}.

In table \ref{tab:R}, several experiments are conducted on the impact of motion scale using a medium-range MAV motion dataset. By varying input resolutions, we evaluated their effect on the performance of the MAV action recognition performance at an input size of $128 \times 128$. Larger-scale event frames improve accuracy by capturing finer spatial details, but they increase computational load and power consumption due to higher-dimensional feature processing. However, larger scales also increase computational load, parameter count, and power consumption. Thus, in real MAV motion signal recognition, an appropriate event frame resolution can be flexibly selected to balance recognition accuracy with computational efficiency.  

\begin{table}[h]
    \scriptsize
    \centering
    \caption{Impact of Input Resolution on Recognition Performance.}
    \label{tab:R}
    \begin{tabular}{cccccc}
    \hline
     Resolution & Accuracy   & ACs(G)        & MACs(G) & Params(M) & Energy(mJ)  \\ \hline
     128x128    & $95.37\%$  & $0.6849$    &$0.424$   &   $1.205$  & $2.567$   \\ 
      64x64     & $92.28\%$  & $0.1636$    &$0.062$  &    $0.418$  & $0.432$  \\
      32x32     & $81.79\%$  &$0.043$      &$0.015$  &    $0.222$  & $0.108$    \\
    \hline
    \end{tabular}
\end{table}

In table \ref{tab:T}, several  experiments on the effect of event timestep $F$ are performed using the same medium-range MAV motion dataset. Timestep $F$ represents the sampling frequency of the time window for MAV motion event streams, encoding them into temporal spike signals for input into our proposed SNN model. By varying $F$ (set to 4, 8, and 16) in ablation studies with equally spaced sampling, we investigated its impact on MAV action recognition accuracy. The results indicate that smaller $F$ significantly reduces computational cost and power consumption but also decreases recognition accuracy. This occurs because a smaller $F$ lowers the sampling frequency of MAV motions, limiting the input spike sequence. Consequently, even with identical model parameters, recognition performance notably declines in MAV action recognition.

\begin{table}[h]
    \scriptsize
    \centering
    \caption{Impact of Time Step $F$ on Recognition Performance.}
    \label{tab:T}
    \begin{tabular}{cccccc}
    \hline
    Time Step & Accuracy      & ACs(G)        & MACs(G)       & Params(M)  & Energy(mJ)\\ \hline
       F=16  &  $95.37\%$    &$0.6849$      &$0.424$  & $1.205$   & $2.567$    \\
       F=8   &  $91.98\%$    & $0.3346$      &$0.1247$  & $1.205$  & $0.875$    \\
       F=4   &  $83.95\%$    &$0.1684$      &$0.0623$   &  $1.205$ & $0.438$    \\
    \hline
    \end{tabular}
\end{table}

Besides, ablation studies are also conducted on the impact of positive and negative events. As shown in the table \ref{tab:PNT}, using only positive (E-PosNet) or negative events (E-NegNet) fails to achieve optimal action recognition performance. The best accuracy is obtained when employing the E-BiPolNet approach (combining positive and negative events for MAV action classification).

\begin{table}[h]
    \scriptsize
    \centering
    \caption{ablation performance analysis in different event polarity. }
    \label{tab:PNT}
    \begin{tabular}{cccc}
    \hline
    Algorithms & Short & Medium & Long   \\ \hline
     E-NegNet  &  $94.12\%$    &$91.49\%$      & $89.05\%$      \\
     E-PosNet  &  $96.32\%$    &$94.33\%$      & $93.43\%$      \\
     E-BiPolNet  &  $96.85\%$    &$95.37\%$      &$94.24\%$    \\
    \hline
    \end{tabular}
\end{table}


\subsection{Visual Communication in Flight Tests}

To validate the feasibility of the proposed motion-based MAV communication framework, flight tests of MAVs are conducted in this section\footnote{A video showcasing the proposed framework and MAV flight tests is provided as supplementary material.}. Similar to other communication systems \cite{kim2024codebook}, we design three different 8-bit communication codes and, based on these codes, used the ``cfclient'' control library from the Crazyflie package to program corresponding MAV action sequences. An event camera is used to observe the MAV's movements and interpret the control information conveyed through the action sequences. The specific meanings of the 8-bit communication codes are shown in Table \ref{tab:flight_signal_encoding}. In each code, the first bit indicates the end of the signal. The bits in between represent, respectively, the flight direction (1 bit), the flight heading angle (3 bits), and the flight distance (2 bits).

\begin{table}[htbp]
\scriptsize
\centering
\caption{MAV Flight Signal Encoding Scheme.}
\label{tab:flight_signal_encoding}
\begin{tabular}{cccp{1.6cm}}
\hline
\textbf{Segment} & \textbf{Bit Range} & \textbf{Meaning} & \textbf{ Example} \\
\hline
Start Flag & Fixed & Start of signal & \texttt{start} \\
Direction & 1 bit (bit 1) & 0 = Forward, 1 = Backward & \texttt{0} (Forward) \\
Heading & 3 bits (bits 2--4) & Heading angle, each step $\alpha^\circ$ & \texttt{000} = $0\times\alpha^\circ$ \\
Distance & 2 bits (bits 5--6) & Flight distance & \texttt{01} = 0.2 m  \\
End Flag & Fixed & End of signal & \texttt{end} \\
\hline
\end{tabular}
\end{table}

\begin{figure}
		\centering 
		\includegraphics[width=0.4\textwidth]{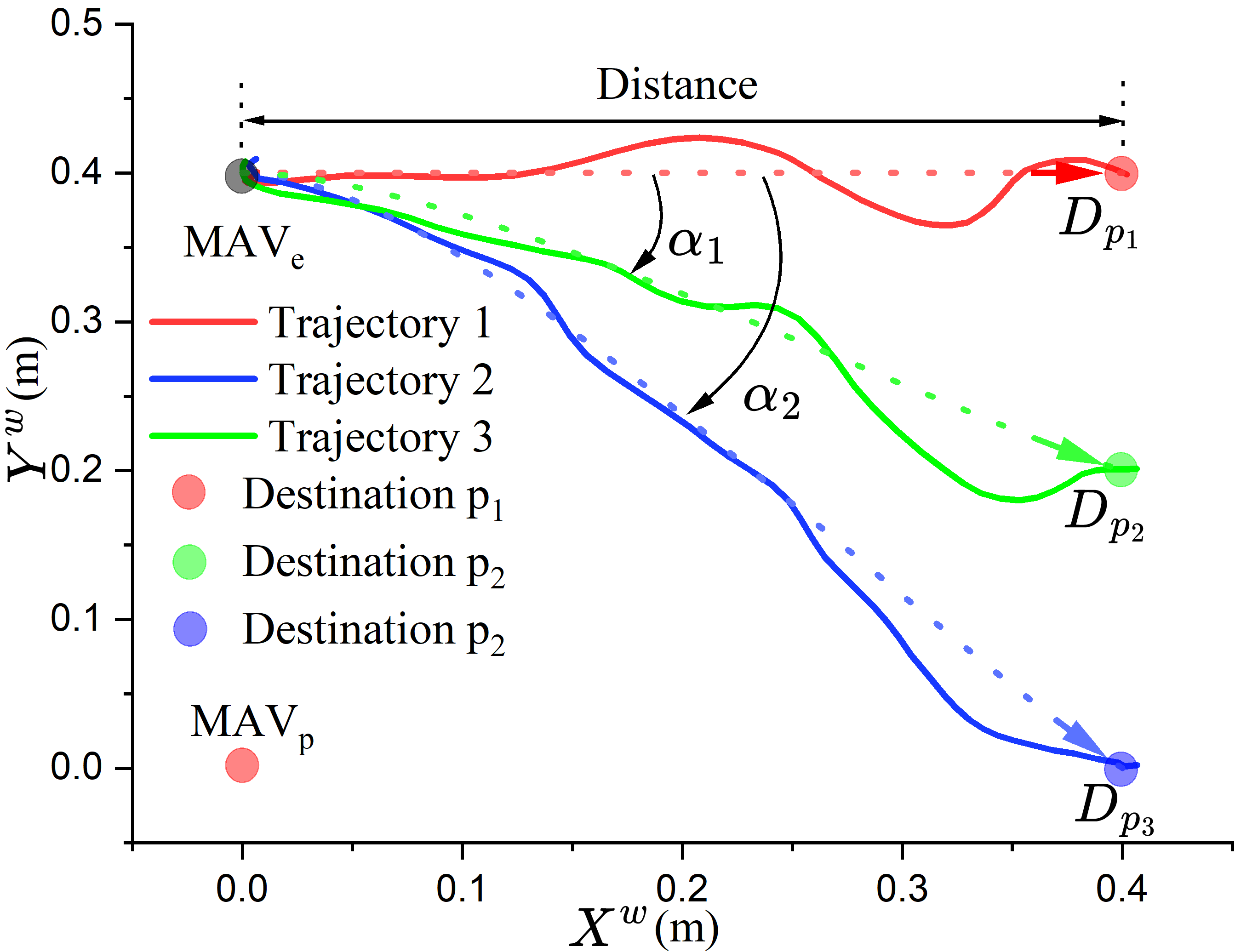}
		\caption{Visual communication in flight tests. The Performer MAV (MAV\textsubscript{p}) transmits the visual message to guide the Executor MAV (MAV\textsubscript{e}) to three different trajectories.} \label{fig:2D_motion}
\end{figure}

\begin{figure}
		\centering 
		\subfigure[MAV action sequence in trajectory 1.]{%
			\resizebox*{0.48\textwidth}{!}{\includegraphics{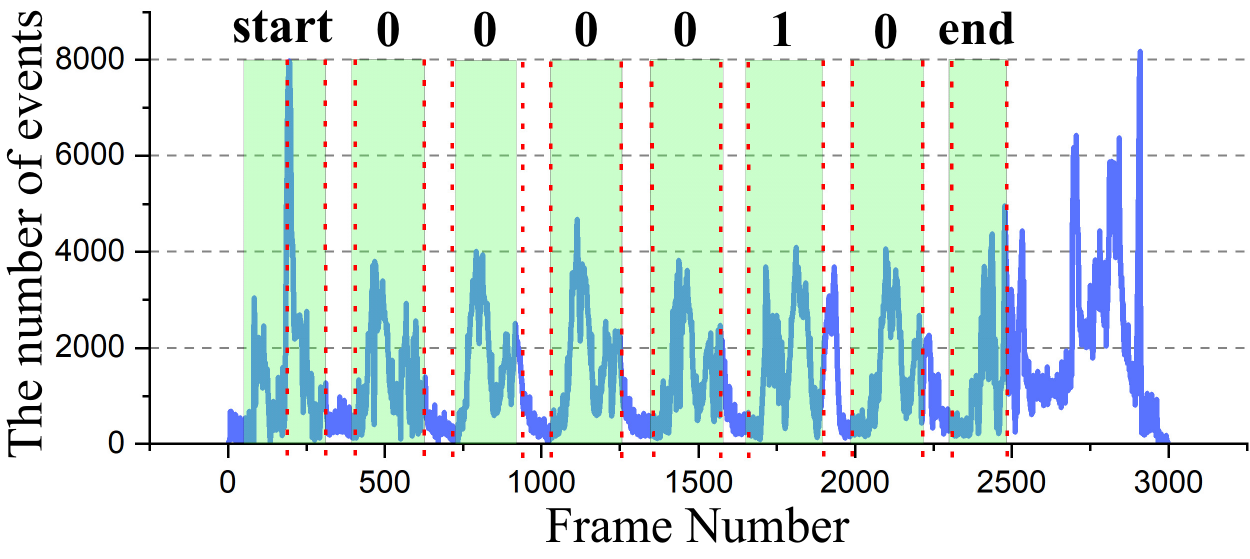}}}
		\subfigure[MAV action sequence in trajectory 2.]{%
			\resizebox*{0.48\textwidth}{!}{\includegraphics{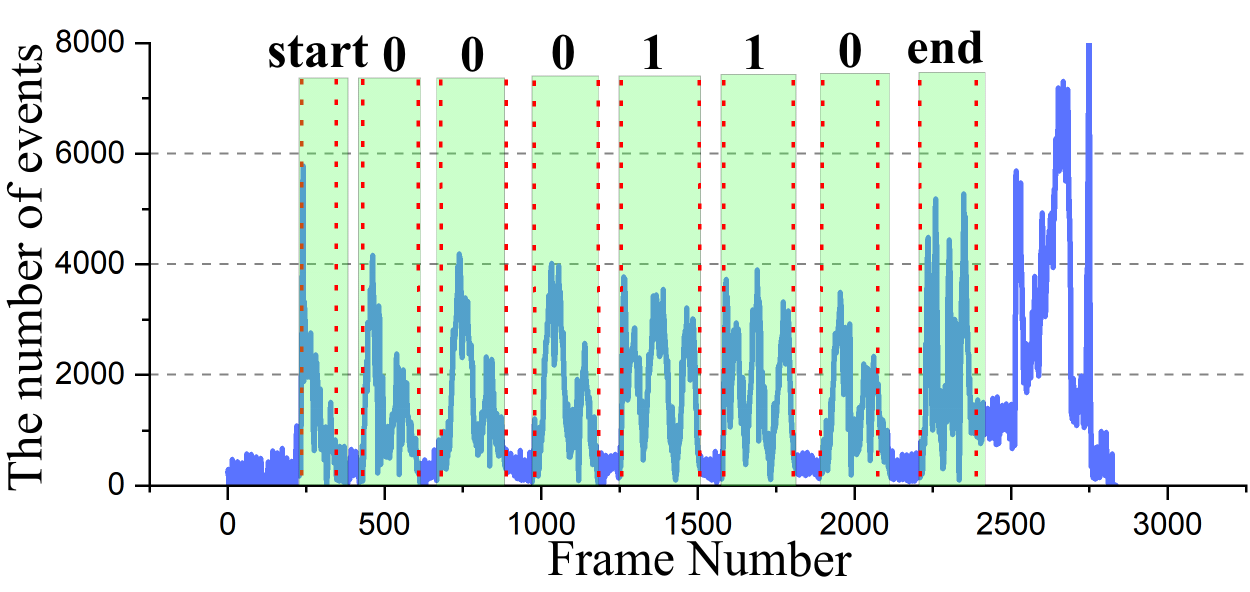}}}
        \subfigure[MAV action sequence in trajectory 3.]{%
			\resizebox*{0.48\textwidth}{!}{\includegraphics{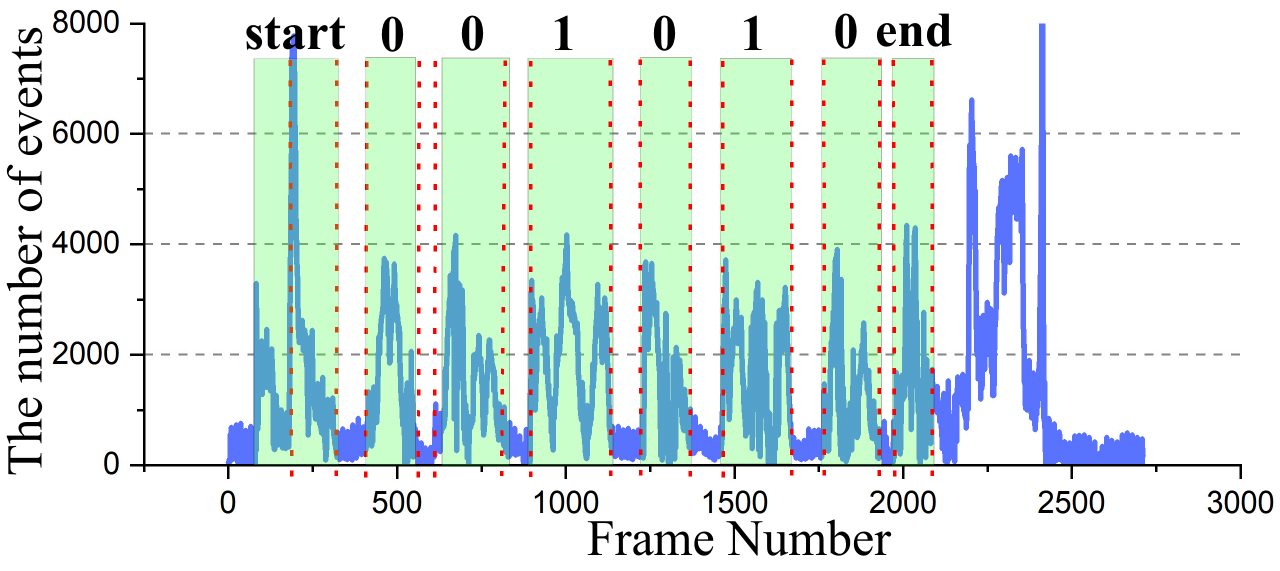}}}            
		\caption{The experimental results of the three flight tests.} 
  \label{fig:UAVDecodingCode}
\end{figure}

To further validate the effectiveness of the proposed visual communication framework, we designed flight communication experiments involving multiple MAVs. In flight tests, two MAVs are deployed: MAV\textsubscript{e} (Executor MAV, which executes navigation commands) and MAV\textsubscript{p} (Performer MAV, which transmits messages via motion). Performer MAV\textsubscript{p} enacts motion patterns to convey messages. Upon receiving the message via the event camera using the proposed method, the signal is decoded and converted into MAV control commands. These commands are then transmitted to the Executor MAV\textsubscript{e}, which subsequently navigates to one of three predefined target destinations, $D_{p_1}$ to $D_{p_3}$. We designed three distinct 8-bit communication codes, as shown in Table \ref{tab:CommunicationCode}, corresponding to the target destinations $D_{p_1}$ to $D_{p_3}$. Ultimately, the proposed method successfully decoded all predefined communication codes, with the decoding results illustrated in Fig. \ref{fig:UAVDecodingCode}. Upon successful code recognition, the Executor MAV\textsubscript{e} executed the control commands and reached the corresponding target destinations. To this end, we recorded the 2D flight trajectories of the Executor MAV\textsubscript{e}, as shown in Fig. \ref{fig:2D_motion}. The results demonstrate that the proposed method not only successfully decodes visual motion information but also accurately controls the Executor MAV\textsubscript{e} to reach the designated positions from $D_{p_1}$ to $D_{p_3}$ .

\begin{table}[h]
    \footnotesize
    \centering
    \caption{The experimental communication code in three flight tests.}
    \label{tab:CommunicationCode}
    \begin{tabular}{cccccc}
    \hline
    Communication code & Begin & Direction & Angle  & Distance    & Stop \\ \hline
     S000010E  &  s    &$0$     &     $000$        & $10$          & e \\
     S000110E  &  s    &$0$     &     $001$      & $10$          & e \\
     S001010E  &  s    &$0$     &     $010$       & $10$          & e \\
    \hline
    \end{tabular}
\end{table}


\section{Conclusion}
This paper presents a novel MAV communication paradigm inspired by non-verbal communication strategies in nature, where motion sequences are used to transmit messages between MAVs. Specifically, MAV information is encoded into binary-coded aerial movements observed by event cameras, enabling successful inter-MAV message decoding and forwarding. By categorizing MAV motions into four primitive types with assigned communication symbols, we establish a fundamental visual communication framework for sharing critical swarm coordination data (e.g., heading, and distance). To achieve efficient visual message transmission, an event-frame-based motion segmentation method using mathematical statistical features is proposed, enabling precise segmentation of MAV action sequences. Furthermore, we design EventMAVNet, a shallow Spiking Neural Network, that outperforms existing event-based action classifiers in both inference speed (1.26 ms / 10 samples) and recognition accuracy (95.37\% in the medium dataset). Experimental results, including a $95.37\%$ recognition accuracy on the medium dataset and successful decoding in flight tests, demonstrate that our event-based motion communication serves as a viable alternative to conventional radio-based methods. Future work will focus on: (1) exploring advanced SNN architectures, such as attention-based or deeper networks, to further enhance MAV action recognition accuracy, (2) expanding MAV action semantic labels by incorporating additional action categories for richer communication. This work represents a significant advancement toward vision-based communication for autonomous MAV swarms.

\section*{Acknowledgment}
The authors would like to acknowledge the support of a research grant for this work. Full details will be provided upon publication.

\ifCLASSOPTIONcaptionsoff
  \newpage
\fi

\bibliographystyle{IEEEtran}
\bibliography{IEEEexample}
\end{document}